\documentclass[sn-nature]{sn-jnl}

\usepackage{graphicx}%
\usepackage{multirow}%
\usepackage{amsmath,amssymb,amsfonts}%
\usepackage{amsthm}%
\usepackage{mathrsfs}%
\usepackage[title]{appendix}%
\usepackage{xcolor}%
\usepackage{textcomp}%
\usepackage{manyfoot}%
\usepackage{booktabs}%
\usepackage{algorithm}%
\usepackage{algorithmicx}%
\usepackage{algpseudocode}%
\usepackage{listings}%
\usepackage{cleveref}
\usepackage{subcaption}
\usepackage{alltt}
\usepackage{svg}
\usepackage{siunitx}

\usepackage{longtable}
\usepackage[inline]{enumitem}
\usepackage{tablefootnote}

\theoremstyle{thmstyleone}%

\theoremstyle{thmstyletwo}%

\theoremstyle{thmstylethree}%

\begin{document}

\title[Article Title]{Assessing how hyperparameters impact Large Language Models' sarcasm detection performance}

\author[1]{\fnm{} \sur{Montgomery Gole}}\email{mgole@torontomu.ca}

\author[1]{\fnm{} \sur{Andriy Miranskyy}}\email{avm@torontomu.ca}

\affil[1]{\orgdiv{Department of Computer Science}, \orgname{Toronto Metropolitan University}, \orgaddress{\street{350 Victoria Street}, \city{Toronto}, \postcode{M5B 2K3}, \state{Ontario}, \country{Canada}}}

\abstract{Sarcasm detection is challenging for both humans and machines. This work explores how model characteristics impact sarcasm detection in OpenAI's GPT, and Meta's Llama-2 models, given their strong natural language understanding, and popularity. We evaluate fine-tuned and zero-shot models across various sizes, releases, and hyperparameters. Experiments were conducted on the political and balanced (pol-bal) portion of the popular Self-Annotated Reddit Corpus (SARC2.0) sarcasm dataset. 
Fine-tuned performance improves monotonically with model size within a model family, while hyperparameter tuning also impacts performance. In the fine-tuning scenario, full precision Llama-2-13b achieves state-of-the-art accuracy and $F_1$-score, both measured at 0.83, comparable to average human performance. In the zero-shot setting, one GPT-4 model achieves competitive performance to prior attempts, yielding an accuracy of 0.70 and an $F_1$-score of 0.75. Furthermore, a model’s performance may increase or decline with each release, highlighting the need to reassess performance after each release.}


\newpage
\maketitle
\section{Declarations}
\begin{itemize}
    \item \textbf{Availability of data and material}: The SARC2.0 dataset can be found in \cite{khodak2017large}. The top performing Llama-2 model's weights are available upon request.

    \item \textbf{Competing interests}: The authors have no competing interests to declare.
    
    \item \textbf{Funding}: This work was partially supported by the Natural Sciences and Engineering Research Council of Canada (grant \# RGPIN-2022-03886). 
    
    \item \textbf{Authors' contributions}: Both authors contributed to the study design and manuscript preparation. Montgomery Gole conducted the material preparation, data collection, coding, experimentation, and analysis. Andriy Miranskyy performed code reviews.
    
    \item \textbf{Acknowledgments}: The authors thank the Department of Computer Science at Toronto Metropolitan University for providing computational resources. The authors thank profusely Dr. Vivian Hu and Dr. Glaucia Melo dos Santos for their insightful discussions regarding this paper.
\end{itemize}

\section{Introduction}\label{sec1}
Both humans and artificial intelligence have difficulty interpreting sarcasm correctly \citep{olkoniemi2016individual, auto_sarcasmdet_survey}. This is especially challenging with textual inputs where body language and speaker intonation are absent \citep{kruger2005egocentrism}. 
Sarcasm detecting agents (i.e., systems that detect sarcasm in texts) are tested on their ability to interpret context when determining whether a textual statement is sarcastic. The development of an accurate sarcasm detection systems holds great potential for improving human-computer interactions, given that sarcasm is widely used in human conversations \citep{olkoniemi2016individual}. To detect sarcasm in text-based social interactions, a model with contextual knowledge and social understanding capabilities is needed. 
There has been extensive work in detecting sarcasm, especially with the Self-Annotated Reddit Corpus (SARC2.0\footnote{SARC2.0 is a the most recent, cleaned version of SARC which offers full context for each observation. Each work mentioned in the body of
\Cref{sec:research_sarc_pol_bal} uses SARC2.0.}) dataset \citep{khodak2017large}. Sarcasm detection models with the highest performance rely on Transformers \citep{potamias2020transformer, fuzzy, commonsense}, recurrent neural networks \citep{bilstm, potamias2020transformer, ilic-etal-2018-deep}, and/or feature engineering \citep{cascade, commonsense}.

Large Language Models (LLMs) have shown effectiveness in natural language understanding tasks~\citep{chen2023chatgpt, zhao2023survey}. However, to our knowledge, there is no comparative study of fine-tuning and zero-shot LLM methods for sarcasm detection, no study of differently versioned ChatGPT models' sarcasm detection abilities, and no study of the effect of LoRA rank, batch size, parameter count, or training epoch amount on fine-tuned Llama-2 models for sarcasm detection. 

Zero-shot testing gives a model a low amount of information about a specific task relative to fine-tuning, and is a basic LLM testing method. This study uses zero-shot testing instead of other inference based testing methods like few-shot testing in order to test LLMs on their latent sarcasm detection abilities.
Furthermore, we use fine-tuning to explore how a pre-trained language model's world knowledge, and language understanding can be transferred to a sarcasm classification task.
In the past, OpenAI has released new versions of models which they claim to be more performant than their older counterparts \citep{openai_devday2023, achiam2023gpt}, we study how differently versioned GPT models detect sarcasm to determine how these claims relate to sarcasm detection.

This work aims to fill this gap by analyzing the performance of GPT and Llama-2 models in detecting sarcasm using the SARC2.0 political, balanced (pol-bal) dataset. 
For the sake of brevity, from hereon, we will refer to this dataset as \textit{pol-bal} dataset. 
Our research questions (RQs) are as follows.
\begin{enumerate}[label=\textbf{RQ\arabic*:},leftmargin=*]
    \item How does model size affect the ability of fine-tuned GPT-3 and Llama-2 models to detect sarcasm?
    \item What are the characteristics of the top-performing zero-shot model under study?
    \item How is zero-shot learning affected by different versions of the same GPT model? 
    \item How is fine-tuned learning affected by different versions of the same GPT model? 
    \item How does LoRA rank, batch size, parameter quantization, and training epoch amount affect fine-tuned learning of Llama-2 models?
    \item How does parameter quantization affect zero-shot learning of Llama-2 chat models?
    
\end{enumerate}
Our key \textbf{contributions} are: 
\begin{enumerate}
    \item We achieve state-of-the-art sarcasm detection performance on the SARC2.0 dataset with Llama-2-13b. 
    \item We analyze how different model sizes (RQ1), versions (RQ2 \& RQ3), and learning methods influence GPT (RQ1, RQ2, RQ3, RQ4) and Llama-2 models' ability to detect sarcasm (RQ5 \& RQ6). 
\end{enumerate}

The rest of this paper is structured as follows.~\Cref{sec:lit_review} presents a review of background literature. ~\Cref{cha:methodology} covers the methodology of our experiments.~\Cref{cha:results} presents the results. These results, along with threats to validity are discussed in ~\Cref{sec:discussion}. Finally,~\Cref{cha:conclusion} concludes the work and poses its potential next step. 

A condensed version of this work~\citep{gole2024sarcasm}, specifically covering the GPT subset of this manuscript appears in the proceedings of and presented at the 34th International Conference on Collaborative Advances in Software and COmputiNg (CASCON 2024).

\section{Literature Review}
\label{sec:lit_review}
The following literature review aims to provide a perspective of how human sarcasm detection has impacted automatic sarcasm detection, an overview of popular sarcasm detection datasets, a description of as well as research on the \textit{pol-bal} dataset, and the implications of Large Language Models (LLMs) on sarcasm detection.

\subsection{From Human Perspectives to automatic Sarcasm Detection}
Sarcasm detection has been explored through neurological, psychological, and linguistic lenses. Social cognition like theory of mind (ToM) is required for humans to detect sarcasm \citep{shamay2005neuroanatomical}. ToM and sarcasm detection relies on the brain's frontal networks \citep{staios2013exploring, UCHIYAMA2006100}, and can be affected by mental health issues that impact auditory functionality like schizophrenia \citep{Kantrowitz_Hoptman_Leitman_Silipo_Javitt_2014}.

Furthermore, a sarcastic statement's context's degree of negativity is positively correlated with the duration of a reader's processing time of said statement's meaning \citep{irony_processing}. 
Linguistically, sarcasm has been categorized into four types: propositional sarcasm, lexical sarcasm, 'like'-prefixed sarcasm, and illocutionary sarcasm \citep{camp2012sarcasm}.

Branching off of these human-centered perspectives, automatic sarcasm detection has roots in computational humor, computational irony, and sentiment analysis. Early works in these fields attempt to generate humor, such as riddles, with rule-based computation \citep{Binsted1994AnIM}, sketch computational models of irony \citep{utsumi1996implicit}, and classify the sentiment of text, as seen in machine learning approaches used to classify the sentiment of feature-engineered movie review data by \citep{Pang+Lee:04a, Pang+Lee:05a, Pang+Lee+Vaithyanathan:02a}. As noted by Ritchie \cite{Ritchie2001CurrentDI} in 2001, a vast foundation of knowledge, and a ``powerful" natural language processing system would be necessary for a computational agent capable to interpret humorous materials.  
Early attempts at automatic sarcasm detection involve testing on speech data \citep{Tepperman2006yeahRS}, and human tagged sarcastic online product reviews via semi-supervised machine learning \citep{Tsur2010ICWSMA}. More recent research on automatic sarcasm detection, as described below, focuses on building datasets for sarcasm detection, and advancing machine learning based sarcasm detection.

\subsection{Modern Sarcsam Detection Datasets}\label{sec:datasets}
Multimodal Sarcam Detection Dataset (MUStARD), developed by Castro et al. \citep{castro-etal-2019-towards}, consists of 690 scenes (50\% sarcastic) from four television shows where a target utterance is to be classified as sarcastic or non-sarcastic.

The iSarcasmEval dataset \citep{abu-farha-etal-2022-semeval} contains 6,135 tweets in English, with 21\% labeled as sarcastic. 

Providing 362 statements (50\% sarcastic), SNARKS \citep{snarks} uses a contrastive minimal-edit distance (MiCE) setup, presenting a binary choice sarcasm detection task. SNARKS omits sarcastic statements requiring factoid-level knowledge as well as the comment thread leading to a sarcastic statement.

\subsection{Dataset under study: SARC2.0 \textit{pol-bal}}\label{sec:sarc_pol_bal}

SARC2.0 is a large dataset containing 1.3 million sarcastic Reddit comments and 533 million total comments from various sub-reddits. The dataset is self-annotated since an observation is labeled as sarcastic or not-sarcastic based on the presence of ``\text{/s}" \citep{khodak2017large}. This self-annotation method is limited by the presence of false-negatives (a sarcastic statement labeled as non-sarcastic due to no ``\text{/s}") which make up 2.0\% of the dataset (discussed further in \Cref{sec:error_analysis}) and false-positives (a non-sarcastic comment where a user included ``\text{/s}") which make up 1.0\% of the dataset. 

Due to its popularity and unique challenges, we utilize SARC2.0's \textit{pol-bal}, balanced dataset with observations from r/politics, containing 13,668 training and 3,406 testing observations.

\subsection{Research on the \textit{pol-bal} dataset}\label{sec:research_sarc_pol_bal}
The following models have been constructed to detect sarcasm in the \textit{pol-bal} dataset, a summary of the models' performance is given in~\Cref{tbl:prior_models} \footnote{Bosselut et al. \citep{commonsense} propose a method for using BERT and COMET-based common sense knowledge, achieving $\approx 0.76$ accuracy and $F_1 \approx 0.76$. They are using a different SARC2.0 subset, preventing direct comparison.

Choi et al. \citep{choi2023llms} benchmark LLMs on social understanding, including sarcasm, using SARC2.0 but not \textit{pol-bal}. DeBERTa-V3 performs best on sarcasm detection but the results are not comparable.

Sharma et al. \citep{fuzzy} apply BERT and fuzzy logic to sarcasm detection but mix SARC2.0 subsets, making comparisons unfeasible.
}.

The ContextuAl SarCasm DEtector (CASCADE) model uses both content and context modeling to classify an r/politics post’s reply as sarcastic. 

This method reaches an accuracy $\approx 0.74$ and $F_1 \approx 0.75$ ~\citep{cascade}.
The authors of SARC2.0 classify their \textit{pol-bal} test set using several models. The best result uses a Bag-of-Bigrams approach and achieves accuracy $\approx 0.77$ \citep{khodak2017large} . 
It is not trivial to detect sarcasm, as stated above. According to \citep{khodak2017large}, five human ``labelers'' attain an average accuracy $\approx 0.83$. A majority vote among the ``labelers'' improves accuracy to $\approx 0.85$.

Pelser \& Murrell \citep{pelser2019deep} use a dense and deeply connected neural model in an attempt to extract low-level features from a sarcastic comment without the inclusion of its respective situational context. 

This method manages to attain accuracy $\approx 0.69$ and $F_1 \approx 0.69$.
 
Ili{\'c} et al. \citep{ilic-etal-2018-deep} attempt to capture more information from a given sarcastic statement. They frame their approach by relying on morpho-syntactic features of a sarcastic statement. 
This method achieves achieves accuracy $\approx 0.79$. 

Potamias et al. \citep{potamias2020transformer} propose a method called RCNN RoBERTa, which uses RoBERTa embeddings fed into a recurrent convolutional neural network to detect sarcasm, achieving accuracy of $\approx 0.79$ and $ F_1 \approx 0.78$. Thus, two methods achieve state-of-the-art accuracy $\approx 0.79$ \citep{ilic-etal-2018-deep, potamias2020transformer}.

\begin{table}[t]
\caption{Performance of models on the SARC2.0 \textit{pol-bal} dataset. A -- indicates cases in which $F_1$ are not reported. Results in \textbf{bold} are state-of-the-art. A * and ** denote a zero-shot, and fine-tuned LLM respectively.}\label{tbl:prior_models}
\begin{tabular}{@{}llrr@{}}
\toprule
Reference                      & Model                       & Acc & $F_1$   \\ \midrule
Our paper           & Llama-2-13b-chat* \tablefootnote{This model's performance is not directly comparable to the others in this table as it ``missed" 1.57\% of classifications. However, it is included for completeness.}          & 0.51   & 0.67 \\
Pelser \& Murrell \cite{pelser2019deep}& dweNet                      & 0.69  & 0.69 \\ 
Gole et al. \cite{gole2024sarcasm}      & GPT-4-0613*  & 0.70  & 0.75 \\
Hazarika et al. \cite{cascade}                 & CASCADE                     & 0.74  & 0.75 \\ 
Khodak et al. \cite{khodak2017large}& Bag-of-Bigrams              & 0.77  &   --   \\ 
Potamias et al. \cite{potamias2020transformer}& RCNN-RoBERTa                & 0.79  & 0.78 \\ 
Ili{\'c} et al. \cite{ilic-etal-2018-deep}& ELMo-BiLSTM                 & 0.79  &  --    \\ 

Gole et al. \cite{gole2024sarcasm}    & GPT-3 175B**                  & 0.81  & 0.81 \\
\textbf{Our paper}        & \textbf{Llama-2-13b}**                 &\textbf{0.835}   & \textbf{0.834} \\
Khodak et al. \cite{khodak2017large}         & Human (Average)             & 0.83  &   --   \\ 
Khodak et al. \cite{khodak2017large}         & Human (Majority)            & 0.85  &  --    \\ \bottomrule
\end{tabular}
\end{table}

\subsection{Research on sarcasm detection with LLMs}
Srivastava et al. \cite{srivastava2023imitation} create the BIG-bench benchmark for natural language understanding tasks with large language models. It details the usage of eight differently-sized OpenAI GPT-3 models from \citep{brown2020language} on 204 natural language tasks in a zero-shot and few-shot manner. In particular, they conduct $[0, 1, 2, 5]$-shot testing on GPT-3 with observations from the SNARKS sarcasm dataset (discussed in~\Cref{sec:datasets}). As they are using a different dataset, their work is complementary to ours. \cite{mu2023navigating} compare the sarcasm-detecting abilities of GPT-3.5-turbo, OpenAssistant, and BERT-large on the iSarcasmEval dataset  (discussed in~\Cref{sec:datasets}). They find that BERT-large is the most performative on this dataset. This work is complementary to ours as it uses a different dataset and a single GPT model. \cite{choi2023llms} create a benchmark for social understanding for LLMs; they group social understanding tasks into five categories, including humor and sarcasm, which includes the SARC2.0 dataset. As discussed in~\Cref{sec:research_sarc_pol_bal}, their work is complementary to ours. \cite{gptsarcasm} use a fine-tuned GPT-3 curie and zero-shot text-davinci-003 models on the MUStARD dataset (discussed in~\Cref{sec:datasets}) and achieve top $F_1=0.77$. This work is complementary to ours as they are using a different dataset. \footnote{Chen et al. \cite{chen2023robust} compare $[0, 1, 3, 5]$-shot LaMDA-PT, $[0, 2, 3, 4, 10, 15, 16]$-shot FLAN, and fine-tuned popular approaches  to $[0, 3, 6, 9]$-shot InstructGPT models (text-davinci-001, text-davinci-002) and the text-davinci-003 GPT-3.5 model on 21 datasets across 9 different NLU tasks not including sarcasm detection. They find that the GPT-3.5 model performs better than the other models in certain tasks like machine reading comprehension and natural language inference, but performs worse than other models in sentiment analysis and relation extraction.}

\section{Methodology}
\label{cha:methodology}
The dataset under study---\textit{pol-bal}---in this project is discussed in \Cref{sec:sarc_pol_bal}. Using this dataset, we study 14 closed source LLMs---GPT models---and 4 open source LLMs--- Meta's Llama 2 models which are shown in~\Cref{models_studies}. We provide overview of the models below.

\begin{table}[t]

\caption{Large Language models under study. * denotes a chat model. The size of the ada, babbage, curie, and davinci models is reported by~\cite{brown2020language}, while the size of each Llama-2 model is reported by ~\cite{touvron2023llama2openfoundation}. The size of the remaining models is unknown. However, it is conjectured that the size of InstructGPT and GPT-3.5 models is similar in magnitude to their predecessors~\cite{zhao2023survey}, while the size of GPT-4 is larger than of their predecessors~\cite{schreiner2023gpt}. }
\label{models_studies}
\begin{tabular}{@{}llrr@{}}
\toprule
Model Family & Version                & \# of Parameters & Release Date  \\ \midrule
GPT-3        & ada                    & 0.4 B      & June 2020        \\
GPT-3        & babbage                & 1.3 B        & June 2020             \\
GPT-3        & curie                  & 6.7 B       & June 2020                \\
GPT-3        & davinci                & 175.0 B      & June 2020            \\
InstructGPT & text-ada-001     & unknown          &  January 2022              \\
InstructGPT & text-babbage-001 & unknown          &    January 2022                       \\
InstructGPT & text-curie-001   & unknown        &          January 2022                \\
GPT-3.5      & text-davinci-003 & unknown          &     January 2022                    \\
GPT-3.5      & gpt-3.5-turbo-0301*         & unknown          & March 2023                  \\
GPT.3.5      & gpt-3.5-turbo-0613*         & unknown          & June 2023                   \\
GPT.3.5      & gpt-3.5-turbo-1106*         & unknown          & November 2023                   \\
GPT-4        & gpt-4-0314*             & unknown          & March 2023                \\
GPT-4        & gpt-4-0613*             & unknown          & June 2023                    \\ 
GPT-4        & gpt-4-1106-preview*             & unknown          & November 2023                    \\ 
Llama-2        & Llama-2-7b             & 7.0 B          & July 2023                    \\ 
Llama-2        & Llama-2-13b             & 13.0 B          & July 2023                    \\ 
Llama-2        & Llama-2-7b-chat*             & 7.0 B          & July 2023                    \\ 
Llama-2        & Llama-2-13b-chat*             & 13.0 B          & July 2023                    \\ 
\bottomrule
\end{tabular}

\end{table}

The GPT and Llama-2 models are generative models pre-trained on a large corpus of text data~\citep{brown2020language, touvron2023llama2openfoundation}. These models are pre-trained to predict the next token in a given document, learning to estimate the conditional probability distribution over its vocabulary given the context; see~\cite{zhao2023survey} for review. With pre-training, these models are equipped with a vast amount of language knowledge and world information, which, in conjunction with their large parameter count, allows them to excel at natural language tasks~\citep{brown2020language}. 

To answer this work's research questions, we initially developed prompts which wrap each observation from the dataset. Subsequently, we fine-tuned and tested Llama-2-7b, Llama-2-13b, GPT-3 and some GPT-3.5 models; zero-shot tested the Llama-2-7b-chat, Llama-2-13b-chat, GPT-3, InstructGPT, GPT-3.5, and GPT-4 models; and finally performed analyses on their results as discussed below.

\subsection{Prompt Development stage} 
This sarcasm classification problem was addressed by creating two prompts\footnote{We tuned these prompts for accuracy and instruction following but do not claim that they are optimal.}: one to wrap input data and a system prompt for zero-shot testing (shown in~\Cref{fig:prompts}). 

We set \texttt{<completion-delimiter>} placeholder to \verb|\\n\\n###\\n\\n| when fine-tuning GPT-3 models. And when performing the rest of the experiments, we set the placeholder to \verb|\n\n###\n\n|. During fine-tuning and at inference, the \texttt{<thread>} placeholder was replaced with a given observation's context (the thread of text leading up to the response), while the \texttt{<reply>} placeholder was replaced with an observation's response. We denote slight differences of prompts per model in \cref{fig:prompts}. We introduced the \texttt{\textbackslash{n}Post\_n:} delimiter between comments in cases where multiple comments were provided within a given observation (as discussed in~\Cref{sec:sarc_pol_bal}). An example of such an observation for the GPT models is given in~\Cref{fig:multi_post_example}, and for the Llama-2-chat models in \Cref{fig:multi_post_example_llama}.

\begin{figure}[tb]
     \begin{subfigure}[b]{0.4\textwidth}
         \centering
         \fbox{\begin{minipage}{\linewidth}
        \begin{alltt}\footnotesize
        Thread: <thread>\textbackslash{n}\textbackslash{n}Reply: <reply><completion-delimiter>
        
        \end{alltt}
        \end{minipage}}

         \caption{Input wrapping prompt.}
         \label{fig:prompt_wo_env}
     \end{subfigure}
     \hfill
     \begin{subfigure}[b]{0.4\textwidth}
         \centering
         \fbox{\begin{minipage}{\linewidth}
            \begin{alltt}\footnotesize
            Classify (each|the) comment thread's (response|reply) as sarcastic with yes or no.\textbackslash{n}
            \end{alltt}
            \end{minipage}}
                     \caption{Zero-shot prompt's prefix.}
         \label{fig:prompt_zero_shot}
     \end{subfigure}
        \caption{Prompts used for fine-tuning and zero-shot testing models. \texttt{<reply>} and \texttt{<thread>} placeholders represent template values that need to be replaced with actual content. (word1\text{\textbar}word2) denotes that \textit{word1} was used in the prompt for GPT experiments, while \textit{word2} was used in the prompt for Llama-2 experiments.}
        \label{fig:prompts}
\end{figure}
\begin{figure}[tb]
     \centering

         \fbox{\begin{minipage}{\linewidth}
            \begin{alltt}\footnotesize
            Classify each comment thread's response as sarcastic with yes or no.\textbackslash{n}Thread: 'Post\_0: '\textbf{\textit{Just finished watching the debate. I love the President!}}'\textbackslash{n}Post\_1: '\textit{\textbf{Agreed! Can't wait for the next event!}}''\textbackslash{n}\textbackslash{n}Reply: '\textit{\textbf{Oh, the suspense is killing me!}}'\textbackslash{n}\textbackslash{n}\#\#\#\textbackslash{n}\textbackslash{n}
            \end{alltt}
            \end{minipage}}
        \caption{A GPT zero-shot test prompt with two comments in a thread.}
        \label{fig:multi_post_example}

\end{figure}
\begin{figure}[tb]
     \centering

         \fbox{\begin{minipage}{\linewidth}
            \begin{alltt}\footnotesize
            Classify the comment thread's reply as sarcastic with yes or no.\textbackslash{n}Thread: 'Post\_0: '\textbf{\textit{Just finished watching the debate. I love the President!}}'\textbackslash{n}Post\_1: '\textit{\textbf{Agreed! Can't wait for the next event!}}''\textbackslash{n}\textbackslash{n}Reply: '\textit{\textbf{Oh, the suspense is killing me!}}'\textbackslash{n}\textbackslash{n}\#\#\#\textbackslash{n}\textbackslash{n}
            \end{alltt}
            \end{minipage}}
        \caption{A Llama-2 zero-shot test prompt with two comments in a thread.}
        \label{fig:multi_post_example_llama}

\end{figure}

\subsection{Fine-tuning stage}\label{sec:method_fine_tune}
Models were fine-tuned in order to explore how pre-trained LLMs' language knowledge can be transferred to the sarcasm detection binary classification task. We conduct fine-tuning for RQ1, RQ4, and RQ5.

\subsubsection{GPT-3}
To test how model size affects a fine-tuned GPT model’s ability to detect sarcasm for RQ1, fine-tuned versions of all the GPT-3 models were created and tested. 

\subsubsection{GPT-3.5}
At the time of experimenting OpenAI did not allow fine-tuning of the GPT-3.5 text-davinci-003 and gpt-3.5-turbo-0314 models~\cite{gpt35_fine_tuning}. 
GPT-3.5-turbo-0613 and gpt-3.5-turbo-1106 models were fine-tuned, but OpenAI does not report their size. Thus, we cannot use them to answer RQ1, but we can use them to answer RQ4.

\subsubsection{GPT-4}
When the experiments were conducted, OpenAI did not allow fine-tuning GPT-4 models. 
We plan to assess the performance of GPT-4 fine-tuned models in the future. 

\subsubsection{Llama-2}
In order to answer RQ1 and RQ5, we fine-tuned Llama-2-7b and Llama-2-13b. Due to computational constraints, none of these models were fully fine-tuned, but instead used the parameter efficient fine-tuning (PEFT) method called low-rank adaptation (LoRA) \citep{hu2021loralowrankadaptationlarge}.dividually.

\subsubsection{Hyperparameters}
The fine-tuning was performed using 4 epochs, a prompt loss weight of 0.01 (applicable only to GPT-3 models), and a batch size of 16. Each model was trained using the prompt shown in~\Cref{fig:prompt_wo_env}. Due to the cost of fine-tuning, we did not perform any hyperparameter grid search with GPT models, but instead used the default/recommended hyperparameters from OpenAI.

The 2 pre-trained Llama models were fine-tuned with different hyperparameter combinations via a grid search of the values seen in~\Cref{tab:hyperparameters}, resulting in 216 fine-tuned Llama-2 models. 

We do not use an exhaustive set of hyperparameters or their values. We chose values that are often used by practitioners for tuning their models. We encourage the community to explore and experiment with other hyperparameters and values.

\begin{table}[ht]
\caption{Hyperparameters used to fine-tune the Llama-2 models}
\centering
\begin{tabular}{@{}l|l@{}}
\toprule
{Hyperparameter} & {Values Tested} \\ \midrule
{LoRA Rank}      & 8, 16, 32, 64                 \\ 
{Batch Size}     & 8, 16, 32             \\ 
{Model Size}     & 7.0 B, 13.0 B    \\ 
{Training Epochs} & 1, 2, 4           \\
{Parameter Precision}   & 4bit, 8bit, 16bit                  \\ \bottomrule
\end{tabular}

\label{tab:hyperparameters}
\end{table}
\subsubsection{Training and validation datasets}
Each fine-tuned model was trained on the \textit{pol-bal} train set, with its desired completion being ``~yes" or ``~no". The training-validation split for each fine-tuned model had a ratio of 75\% for training and 25\% for validations (shuffled at random once and then reused for each fine-tuning session).

\subsection{Zero-shot stage}\label{sec:method_zero_shot}
Zero-shot testing methods were used to explore how the pre-trained LLMs under study can detect sarcasm with its only information about \textit{pol-bal} being one observation. This testing method is used for RQ2, RQ3, and RQ6. For the GPT family, all models in \cref{models_studies} were subjected to zero-shot testing, while within the Llama-2 family, we only tested the chat models. Zero-shot testing was conducted with the \textit{pol-bal} test set.

The output of the models with and without logit bias is compared. Also reported are the number of observations that did not return ``yes'' or ``no'' keywords. Furthermore, the performance of the Llama-2 chat models with different parameter precisions is explored via RQ6.

\subsection{Data analysis stage}\label{sec:method_data_analysis}

\subsubsection{Performance Analysis}
Accuracy, $F_1$-score, and Matthews Correlation Coefficient (MCC) metrics are used to measure the performance of classification methods. These metrics we used (not including MCC) are the same as those used by other researchers studying this dataset (see~\Cref{sec:sarc_pol_bal} for details). To establish a baseline, we use the naive ZeroR classifier, which labels every test observation as sarcastic. Given that the dataset is balanced, ZeroR classifier accuracy $= 0.5$ and $F_1 \approx 0.67$. As described in~\Cref{sec:method_zero_shot}, some models cannot label some observations. Such observations are removed from the list of observations used to calculate accuracy and $F_1$-score. The exact version of McNemar's test~\cite{mcnemar1947note} is used to determine if there was a statistically significant difference between the classification methods' performance, as recommended by~\cite{dietterich1998approximate}. Multiple linear and random forest \citep{breiman2001random} regressions were used to determine the answers to RQ1 and RQ5 regarding fine-tuned Llama-2 models. As the performance data for our Llama-2 models did not follow a normal distribution, a Wilcoxon signed-rank test \cite{wilcoxonsigned} was performed to assess the statistical significance of differences in model performance with respect to each hyperparameter.

\subsubsection{Data Preprocessing and Preparation}
\label{subsub:datapre}
Prior to performing analysis, univariate outliers were identified and removed for each target variable. Outliers were detected using the interquartile range method, where a data point is excluded from analysis if it is outside the range of 1.5 times IQR (interquartile range) below the first quartile or 1.5 times IQR above the third quartile.
\subsubsection{Regression Analysis of Hyperparameter Effects}
\label{subsub:open_src_regressn}
We employed multiple linear regression to capture linear relationships between hyperparameters and the performance metrics. 

A random forest regression was utilized to capture non-linear interactions between the hyperparameters and the performance metrics variables. 

\subsection{Error Analysis stage}
An analysis of which observations a model classifies incorrectly can give insight into how the dataset and model interact. We conduct an error analysis of the top performing Llama-2 model with SHAP \citep{shap} as a part of our discussion in \Cref{sec:error_analysis}.

\section{Results}
\label{cha:results}

Below are the results of the experiments described in~\Cref{cha:methodology}. Sections~\ref{sec:rq_size}~--~\ref{sec:rq6llamazero} present the results as answers to our six research questions (defined in \Cref{sec1}). \Cref{sec:discussion} discusses the results, while \Cref{sec:threats} identifies possible threats to their validity.

\subsection{RQ1: How does model size affect the ability of fine-tuned GPT-3 and Llama-2 models to detect sarcasm?}
\label{sec:rq_size}
    
\begin{figure}[H]
    \centering
    \includegraphics[width=0.7\textwidth]{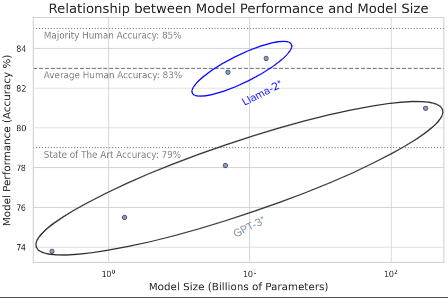}
    \caption{fine-tune results, comparing accuracy to parameter count. The top performing Llama-2-7b and Llama-2-13b models are shown, along with each fine-tuned GPT-3 model.}
    \label{fig:fine_tune_graph}
\end{figure}

The \textbf{answer to RQ1} is as follows. In~\Cref{tbl:results-fine-tuned}, the accuracy and $F_1$-scores of all fine-tuned models with respect to each model group tested increased monotonically with model size. A more succinct version of this table can be found in \Cref{fig:fine_tune_graph}.  Below, we provide a rationale for the answer.

\subsubsection{GPT-3}
\begin{figure}[H]
\centering
\includegraphics[width=0.8\textwidth]{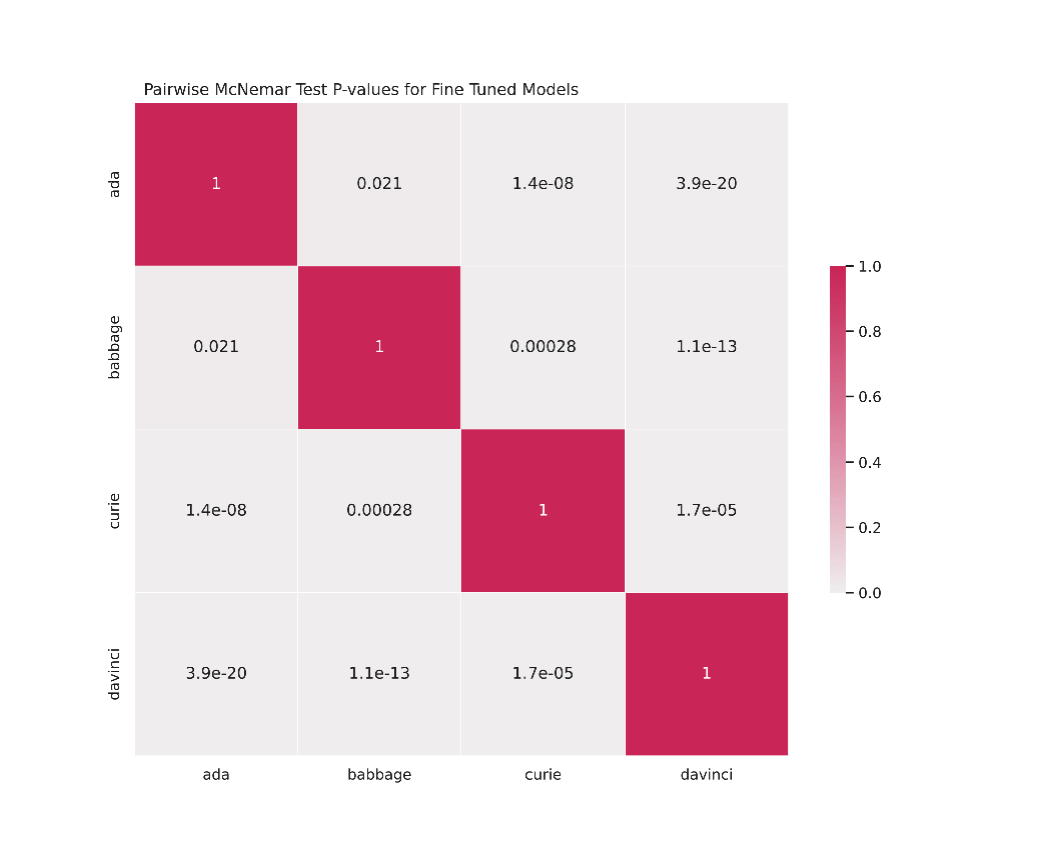}
\caption{Pairwise McNemar's test of fine-tuned base GPT-3 models.}
\label{results-mcnemar-ft-noenv}
\end{figure}

To ensure that the difference in GPT model performance was statistically significant, we performed a pairwise analysis using McNemar's test for all fine-tuned GPT models. According to the test, the differences between all the models are statistically significant ($p < 0.05$), as seen in~\Cref{results-mcnemar-ft-noenv}.

\subsubsection{Llama-2}
To ensure a statistically significant difference in performance for Llama-2 models, a linear regression based on accuracy (dependent) and model size (independent), and one-tailed Wilcoxon signed-rank test was performed, Both tests used our data without outliers (see~\Cref{subsub:datapre} for details on detecting outliers). \Cref{tbl:fine_tune_model_size_regression_accuracy} shows that model size has a significant, positive effect on accuracy. Our Wilcoxon signed-rank test was performed with the \textit{scipy} Python library based on our dataset of testing accuracies for 7 billion parameter experiments and 13 billion parameter experiments with outliers removed. 
With $H_0$ defined as \textit{no significant difference between the two samples, with the accuracy of the experiments using  13 billion parameters models not exceeding that of the experiments using 7 billion parameter models}, the test resulted in a $p$-value \num{< 1e-10}, prompting us to reject $H_0$ and conclude that there is a significant difference between the test accuracy of our 7 billion parameter and 13 billion parameter models, and  that fine-tuned 13 billion parameter Llama-2 models garner a significantly higher accuracy than the 7 billion parameter models. 
Given the significant, positive impact of model size on accuracy, and the observed differences across Llama-2 model sizes, we conclude that model size has a positive impact on a Llama-2 model's ability to detect sarcasm.

\begin{table}[htb]
\caption{Estimated regression parameters, standard errors, t-value, and $p$-values of Llama-2 model size w.r.t test accuracy. The dataset size was reduced from 216 to 201 after removing outliers. Adjusted $R^2=0.1881$, $p$-value\num{=7.569e-11}.}
\label{tbl:fine_tune_model_size_regression_accuracy}
\centering

\begin{tabular}{@{}lSSSS@{}}
\toprule
 & {Estimate} & {Std. error} & {$t$-value} & {$p$-value} \\ \midrule
(Intercept) & 0.8098636 & 0.0015291 & 529.620 & < 2e-16 \\
model\_size & 0.0009983 & 0.0001451 & 6.881 & 7.57e-11 \\ \bottomrule
\end{tabular}

\end{table}

\begin{table}[tb]
\caption{Classification results of \textit{pol-bal} for fine-tuned models. Best results are in \textbf{bold}. The Missed column indicates the percentage of test observations with no labels returned. }\label{tbl:results-fine-tuned}
\centering

\begin{tabular}{@{}llrrrr@{}}
\toprule
Model Family       & Model              & Parameters    & \multicolumn{3}{c}{Performance} \\ \cmidrule(lr){4-6}
   &                                    & count         & Acc           & $F_1$         & Missed (\%) \\ \midrule
\multirow{4}{*}{GPT-3}   & ada                & 0.4 B   & 0.738 & 0.737 & 0.00\% \\
                         & babbage            & 1.3 B   & 0.755 & 0.755 & 0.00\% \\
                         & curie              & 6.7 B    & 0.781 & 0.784 & 0.00\% \\
                         & davinci            & 175.0 B & 0.810 & 0.808 & 0.00\% \\ \midrule
\multirow{2}{*}{GPT-3.5} & gpt-3.5-turbo-0613 & unknown & 0.776 & 0.786 & 0.00\% \\
                         & gpt-3.5-turbo-1106 & unknown & 0.781 & 0.803 & 0.00\% \\ \midrule
\multirow{2}{*}{Llama-2} & Llama-2-7b & 7.0 B & 0.829 & 0.832 & 0.00\% \\
                         & Llama-2-13b & 13.0 B & \textbf{0.835} & \textbf{0.834} & 0.00\% \\ \bottomrule
\end{tabular}
\end{table}

\subsection{RQ2: What are the characteristics of the top-performing zero-shot model under study?}\label{sec:rq_zero_shot}
~\Cref{tbl:results-zero-shot} shows that the GPT-3, InstructGPT, Llama-2-chat and GPT-3.5 text-davinci-003 models perform worse than the ZeroR classifier (described in~\Cref{sec:method_data_analysis}). Although logit bias reduces the count of missing observations, low accuracy and $F_1$-scores indicate that the models cannot differentiate between sarcastic and non-sarcastic comments. Even for top-performing models (e.g., GPT-4), the addition of logit bias leads to performance degradation. Logit bias is not available with Llama-2-chat models, therefore, logit bias performance was not measured for these models.

Some versions of GPT-3.5-turbo and GPT-4 models perform better than the ZeroR baseline. However, except for GPT-4 gpt-4-0613 and gpt-4-1106-preview, their performance is lower than those of the simpler models shown in~\Cref{tbl:prior_models}.

\begin{figure}[htb]
    \centering
    \includegraphics[width=0.7\textwidth]{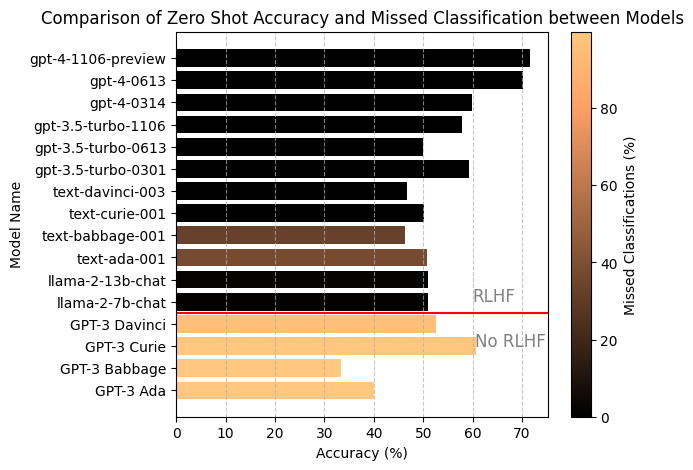}
    \caption{Zero shot results for unbiased models. Models above solid red line were fine-tuned with Reinforcement Learning from Human Feedback. A darker coloured bar represents less missed classifications.}
    \label{fig:zero_shot_non_biased}
\end{figure}

\Cref{fig:zero_shot_non_biased} visualizes a comparison of zero shot accuracy and missed classifications between non-logit-biased models under study. Models fine-tuned with Reinforcement Learning from Human Feedback (RLHF) outperform models without RLHF.

The zero-shot models ``challenge'' is won by GPT-4 GPT-4-0613 model, achieving accuracy~$\approx 0.70$ and $F_1 \approx 0.75$. This model had no missing observations.
Note that gpt-4-1106-preview achieves higher accuracy~$\approx 0.72$ and lower $F_1 \approx 0.74$, but this model answered ``yes" or ``no" only to $99.91$\% of observations.

Based on this analysis, the \textbf{answer to RQ2} is as follows. In the \textit{pol-bal} dataset, only the most sophisticated GPT model (i.e., GPT-4 gpt-4-0613) can detect sarcasm competitively using the zero-shot approach. However,~\Cref{tbl:prior_models} shows that gpt-4-0613 model performs poorly in comparison with prior simpler models, placing it second-last among them.

\begin{table*}[tb]
\centering
\resizebox{\textwidth}{!}{
\begin{tabular}{@{}ll|rrr|rrr@{}}
\toprule
Model Family       & Model              & \multicolumn{3}{c|}{Performance (w/o bias)} & \multicolumn{3}{c}{Performance  (w bias)} \\ \cmidrule(lr){3-5} \cmidrule(lr){6-8}
   &                                    & Acc   & $F_1$ & Missed (\%) &  Acc   & $F_1$ & Missed (\%) \\ \midrule
\multirow{4}{*}{GPT-3}       & ada                & 0.400 & 0.000 & 99.85 & 0.500 & 0.666 & 0.00 \\
                             & babbage            & 0.333 & 0.000 & 99.82 & 0.500 & 0.000 & 0.00 \\
                             & curie              & 0.607 & 0.645 & 99.18 & 0.500 & 0.000 & 0.00 \\
                             & davinci            & 0.526 & 0.076 & 95.48 & 0.509 & 0.658 & 0.00 \\ \midrule
\multirow{3}{*}{InstructGPT} & text-ada-001       & 0.508 & 0.646 & 37.52 & 0.500 & 0.604 & 0.00 \\
                             & text-babbage-001   & 0.463 & 0.541 & 32.47 & 0.484 & 0.503 & 0.00 \\
                             & text-curie-001     & 0.501 & 0.667 & 0.15  & 0.500 & 0.667 & 0.00 \\ \midrule
\multirow{4}{*}{GPT-3.5}     & text-davinci-003   & 0.467 & 0.266 & 0.00  & 0.479 & 0.404 & 0.00 \\
                             & gpt-3.5-turbo-0301 & 0.592 & 0.653 & 0.03  & 0.587 & 0.656 & 0.00 \\
                             & gpt-3.5-turbo-0613 & 0.500 & 0.002 & 0.00  & 0.499 & 0.032 & 0.00 \\
                             & gpt-3.5-turbo-1106 & 0.578 & 0.613 & 0.00  & 0.504 & 0.581 & 0.00 \\ \midrule
\multirow{6}{*}{Llama-2-chat}& llama-2-7b (16 bit)   & 0.510 & 0.658 & 1.41  & --- & --- & --- \\
                             & llama-2-7b (8 bit) & 0.504 & 0.658 & 1.41  & --- & --- & --- \\
                             & llama-2-7b (4 bit) & 0.486 & 0.593 & 2.38  & --- & --- & ---  \\
                             & llama-2-13b (16 bit) & 0.509 & 0.666 & 1.57  & --- & --- & ---  \\
                             & llama-2-13b (8 bit) & 0.518 & 0.668 & 0.85   & --- & --- & --- \\
                             & llama-2-13b (4 bit) & 0.578 & 0.613 & 0.18   & --- & --- & --- \\ \midrule
\multirow{3}{*}{GPT-4}       & gpt-4-0314         & 0.599 & 0.710 & 0.00  & 0.601 & 0.712 & 0.00 \\
                             & gpt-4-0613         & \textbf{0.701} & \textbf{0.748} & \textbf{0.00}  & 0.680 & 0.738 & 0.00 \\
                             & gpt-4-1106-preview & 0.717 & 0.739 & 0.03  & 0.694 & 0.744 & 0.00 \\ \bottomrule
\end{tabular}
}

\caption{Classification results of \textit{pol-bal} for zero-shot models. Best results are in \textbf{bold}. The Missed column indicates the percentage of test observations with no labels returned by the model.}
\label{tbl:results-zero-shot}
\end{table*}

\subsection{RQ3: How is zero-shot learning affected by different versions of the same GPT model? }\label{sec:rq3}
\label{sec:rq_zero_version}
In our work, we have two models with three versions, namely 
\begin{enumerate}
    \item GPT-3.5-turbo released in March (0301), June (0614), and November (1106) of 2023 and
    \item GPT-4 released in March (0314), June (0614), and November (1106-preview) of 2023.
\end{enumerate}

\begin{figure}[H]
\centering
\includegraphics[width=0.6\textwidth]{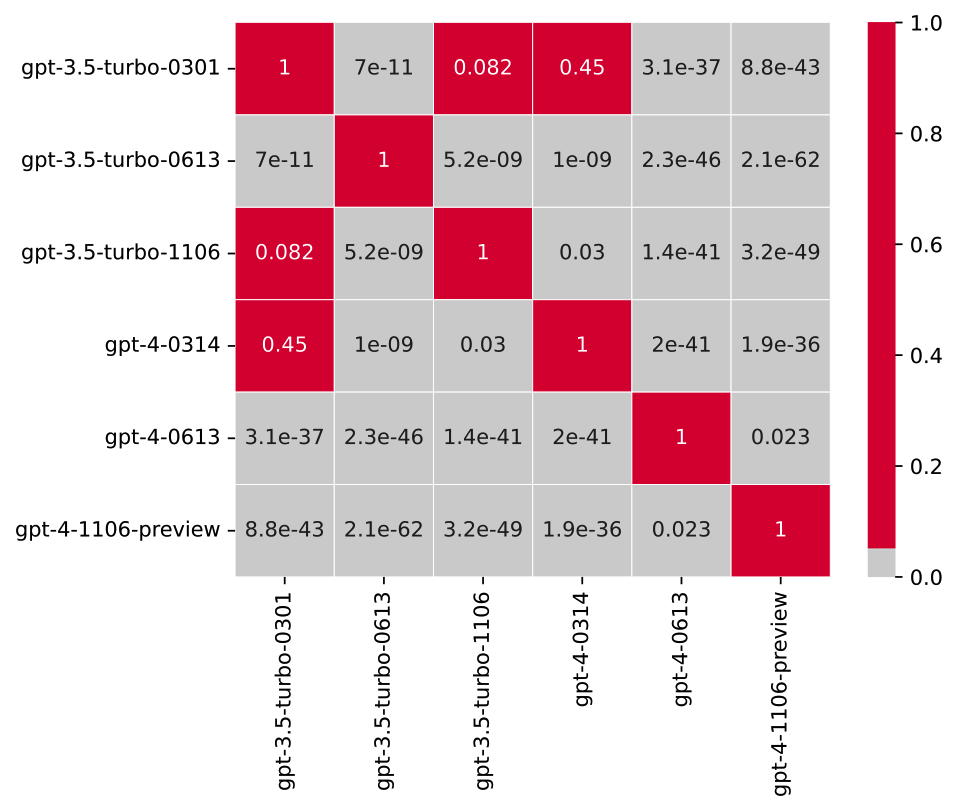}
\caption{Pairwise McNemar's test of ChatGPT models without logit bias.}
\label{results-mcnemar-zeroshot-chat-nodomain-nologit}
\end{figure}

McNemar's test $p$-values for these models are seen in in~\Cref{results-mcnemar-zeroshot-chat-nodomain-nologit}. According to the results, the difference between most releases is statistically significant ($p < 0.05$). 

However, there are some exceptions. In the absence of bias, according to McNemar's test, gpt-3.5-turbo-0301 yields results similar to gpt-3.5-turbo-1106 and gpt-4-0314. 
With bias, gpt-3.5-turbo-0301 is similar to gpt-4-0314, and gpt-3.5-turbo-0613 is similar to gpt-3.5-turbo-1106.
However, the differences in accuracy and $F_1$ values produced by these models may suggest that this similarity is an artifact of statistics.

As shown in~\Cref{tbl:results-zero-shot}, with bias, accuracy and $F_1$ increased monotonically for GPT-4. Every new GPT-3.5-turbo model has not reached the performance of its first release (gpt-3.5-turbo-0301). A bias-free comparison is more challenging because some models lack observations, but we can see that the latest models aren't the best.

In other words, \textbf{the answer to RQ3} is as follows. The GPT model's ability to detect sarcasm may decline or improve with new releases, as was observed for other tasks by~\cite{chen2023chatgpt}.

\subsection{RQ4: How is fine-tuned learning affected by different versions of the same GPT model? }\label{sec:fine_tuned_version}
\begin{figure}[H]
\centering
\includegraphics[width=0.5\textwidth]{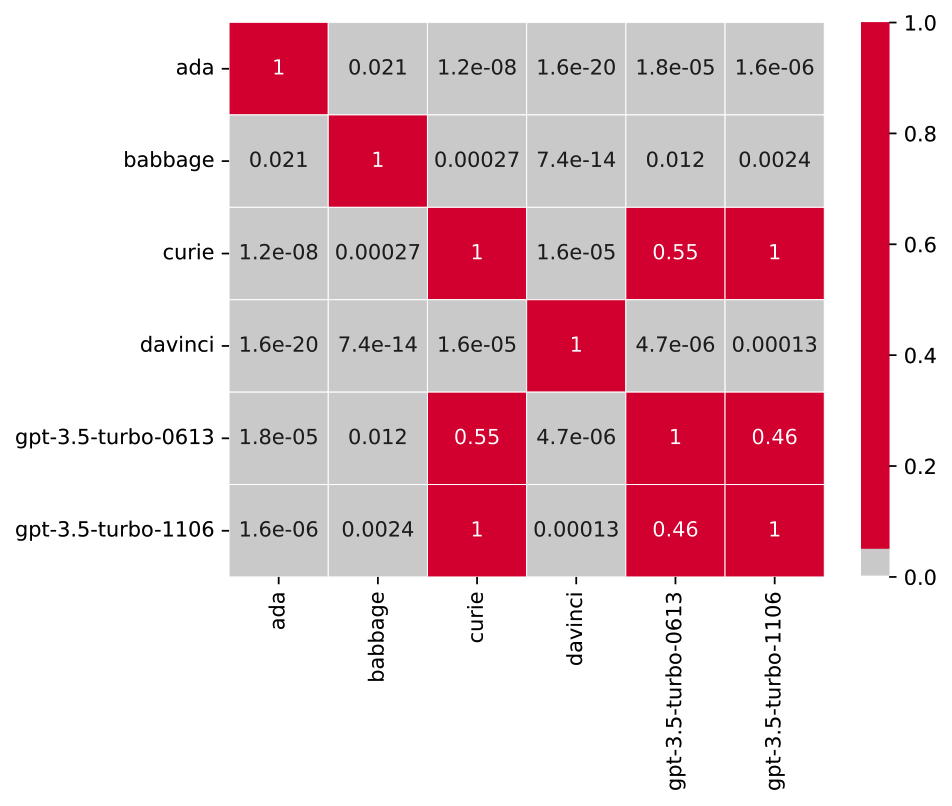}
\caption{Pairwise McNemar's test of fine-tuned differently versioned GPT-3.5 and GPT-3 models.}
\label{results-mcnemar-ft-version}
\end{figure}

According to~\Cref{tbl:results-fine-tuned}, both GPT-3.5 models (gpt-3.5-turbo-0613 and gpt-3.5-turbo-1106) perform similarly, although a newer model exhibits slightly better performance, with accuracy rising from 0.776 to 0.781 and $F_1$-score increasing from 0.786 to 0.803. According to McNemar's test (shown in~\Cref{results-mcnemar-ft-version}), this difference is not statistically significant.

Fine-tuned GPT-3.5 models perform worse than the GPT-3 davinci model. McNemar's test reveals that the performance of both GPT-3.5 models is comparable (from a statistical perspective) to that of GPT-3 curie. Since OpenAI does not report the size and architecture of GPT-3.5 models, it is difficult to draw strong conclusions from this observation.

\textbf{The answer to RQ4} is as follows. Versioning of fine-tuned models does not have a significant impact on the model's performance. Newer versions of the models may change the answer in the future.

\subsection{RQ5: How does LoRA rank, batch size, parameter quantization, and training epoch amount affect fine-tuned learning of Llama-2 models?} \label{sec:llama_hyperparams_fine_tuned}

Performing a linear regression attempting to fit the dependent variable--test accuracy--to our independent variables--LoRA rank, batch size, parameter quantization, and training epoch amount--for our fine-tuned Llama-2 models' results in an adjusted $R^2$ value of 0.2133, and $p$-value $ < 0.05$. The details of the statistics are given in~\Cref{tbl:fine_tune_regression_accuracy}. 

Our analysis shows that~---~relative to our specific task and dataset~---~a fine-tuned Llama-2 model's training epoch amount has the most significant effect on its accuracy. The epoch count has a negative correlation with the test accuracy, while the model size is positively correlated with the test accuracy (as seen in \Cref{sec:rq_size}). Batch size, parameter precision (in bits), and LoRA rank do not have a significant effect on the test accuracy. 

\begin{figure}[htb]
    \centering
    \includegraphics[width=1.1\textwidth]{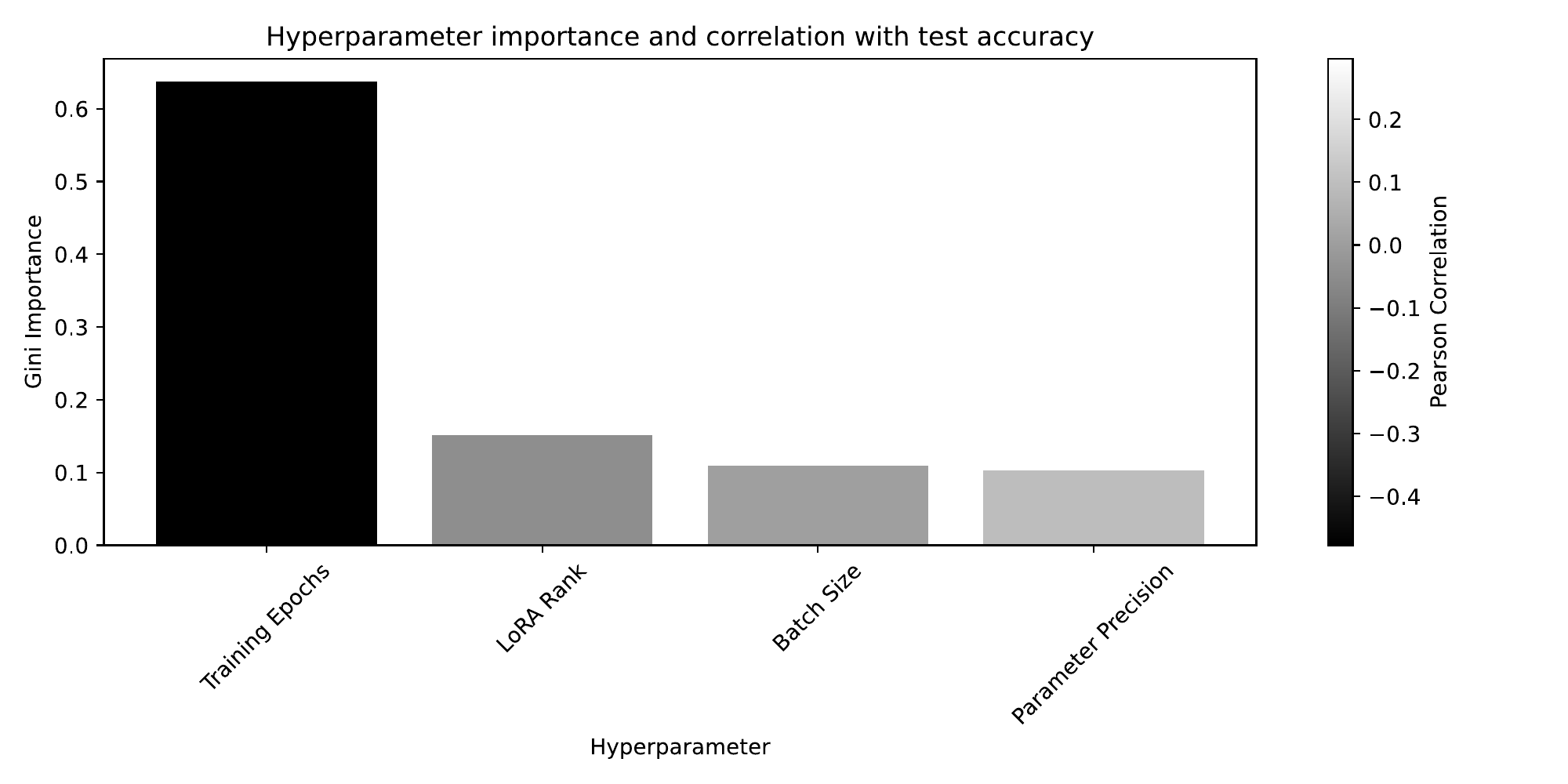}
    \caption{Hyperparameter importance from the optimal random forest model, with hyperparameter Pearson correlation coefficient to accuracy indicated by color.}
    \label{fig:rf_feature_importances}
\end{figure}

When performing a random forest regression with 61 estimators (see estimator amount selection process in~\Cref{subsub:open_src_regressn}), as seen in~\Cref{fig:rf_feature_importances}, we obtained similar results to the linear regression described above. Epoch count is the most important hyperparameter, while LoRA rank, batch size, and parameter precision (in bits) are the least important. Furthermore, we calculated Pearson's correlation coefficients for each hyperparameter with respect to accuracy, showing that the LoRA rank, parameter precision, and batch size have $\approx 0$ correlation with accuracy, while training epoch count has a negative correlation.

\begin{table}[]
\caption{Hyperparameter importance from the optimal random forest model, with hyperparameter Pearson correlation coefficient to accuracy.}
\label{tbl:rf_feature_importance}
\centering
\begin{tabular}{@{}lSS@{}}
\toprule
{Hyperparameter} & {Gini Factor Importance} & {Correlation with Accuracy} \\ \midrule
Batch Size & 0.110 & 0.097  \\
Lora Rank & 0.151 & -0.048 \\ 
Parameter Precision & 0.102 & 0.006 \\ 
Epoch Count & 0.640 & -0.480 \\ \bottomrule 
\end{tabular}
\end{table}
Further statistical analysis via pair-wise, right-sided Wilcoxon Signed-Rank tests show that the different hyperparameter values tested (for a list of values see~\Cref{tab:hyperparameters}) result in statistically significant differences. Specifically:
\begin{itemize}
    \item The mean accuracy for runs with batch size of 16 results is significantly higher accuracy than runs with batch size of 8.
    \item The mean accuracy for runs with parameter precision of 16 bits is significantly greater than precision of 4 and 8.
    \item The mean accuracy for runs with LoRA rank of 64 is significantly less than LoRA rank of 32 and 16.
    \item The mean accuracy for runs with an epoch count of 2 is significantly greater than runs with epoch count of 1 and 4, while the mean accuracy for epoch count of 4 is significantly less than 2 and 1.
\end{itemize}
Parameter precision, batch size, and LoRA rank do not differ significantly with respect to test accuracy ($p > 0.05$). However, it can be seen in~\Cref{fig:utest_epoch_count} that a lower epoch count results in a significantly higher test accuracy for all models tested. Specifically, models with an epoch count of 2 resulted in significantly better performance than any other epoch count value tested.

In summary, \textbf{the answer to RQ5} is as follows. LoRA rank, parameter precision, and batch size have an insignificant effect on fine-tuned learning of the Llama-2 model we tested. The amount of training epochs has a negative effect, which means that as the epoch amounts increase, accuracy on \textit{pol-bal} dataset decreases (i.e., the model starts to overtrain quickly).

\begin{table}[tb]
\caption{Estimated regression parameters, standard errors, t-value, and $p$-values of Llama-2 hyperparameters w.r.t test accuracy. The dataset size was reduced from 216 to 201 after removing outliers (see~\Cref{subsub:datapre} for details). Adjusted $R^2=0.2133$, $p$-value\num{< 2.2e-16}.}
\label{tbl:fine_tune_regression_accuracy}
\centering
\begin{tabular}{@{}lSSSS@{}}
\toprule
 & {Estimate} & {Std. error} & {$t$-value} & {$p$-value} \\ \midrule
(Intercept)               & 8.258e-01 & 1.418e-03 & 582.321 & < 2e-16 \\
Parameter Quantization                     & 5.097e-05 & 3.491e-05 & 1.460 & 0.146 \\
Batch Size  & 3.649e-06 & 4.295e-05 & 0.085 & 0.932 \\
LoRA Rank   & -2.227e-05 & 2.035e-05 & -1.094 & 0.275 \\
Epoch count               & -2.586e-03 & 3.459e-04 & -7.475 & 2.51e-12 \\ \bottomrule

\end{tabular}
\end{table}

\subsection{RQ6: How does parameter quantization affect zero-shot learning of the Llama-2 chat models tested?} \label{sec:rq6llamazero}

We conducted McNemar's test to determine a difference between three parameter precision values (4 bit, 8 bit, 16 bit) for each model size. As seen in \Cref{fig:mcnemar-llama-chat-7b}, the 7 billion parameter Llama-2-chat model with 4 bit parameter precision results in significantly different performance than the other parameter precisions tested. Furthermore, \Cref{fig:mcnemar-llama-chat-7b} shows that the 13 billion parameter Llama-2-chat model with 8 bit parameter precision performs significantly differently from its 4 bit and 16 bit counterparts. ~\Cref{tbl:results-zero-shot} there is too low of a difference in performance metrics between these tests to comment on the correlation between parameter precision and accuracy, $F_1$ score, and missed classifications.

Thus, \textbf{the answer to RQ6 is} that, compared to 8-bit, and 16-bit, 4-bit parameter precision results in significantly different performance for the 7 billion-parameter model, while 8-bit precision results in significantly different performance to other precisions tested for the 13 billion parameter model.

\section{Discussion}
\label{sec:discussion}

\subsection{Miscellanea}

Regarding RQ1 (see \Cref{sec:rq_size}), while fine-tuned performance was found to increase monotonically for GPT-3 and Llama-2 models respectively, performance does not increase monotonically when comparing all of the models together because the Llama-2 models (7b and 13b parameters) outperform the largest GPT-3 model (175b parameters) tested. Due to the black box nature of the GPT-3 models and their fine-tuning algorithm, we cannot be sure why this is the case, however, this could be attributed to the fact that the Llama-2 models underwent hyperparameter tuning, while the GPT-3 models did not. 

Regarding our results from RQ2 in \Cref{sec:rq_zero_shot}, we find that the top performing zero shot model is GPT-4, with GPT-4-0613 and GPT-4-1106-preview achieving accuracies comparable to previous attempts seen in~\Cref{tbl:prior_models}. Interestingly, models which are fine-tuned with Reinforcement Learning from Human Feedback miss far less classifications than models fine-tuned without RLHF. This probably occurs because RLHF is meant to align a model with human instructions. 

Regarding our results from \Cref{sec:llama_hyperparams_fine_tuned}, as seen in \Cref{tbl:results-fine-tuned}, the top performing model on \textit{pol-bal} is Llama-2-13b fine-tuned with batch size 8, LoRA rank 16, 2 epochs, and full precision. This model is limited to outputting 1 or 0, making it impractical to be used by an end-user in a chat-bot setting. These results show that low compute values \textit{may} be acceptable for fine-tuning jobs for difficult NLU tasks in practice with LLMs.

\subsection{Top Performer Error Analysis}
\label{sec:error_analysis}
\Cref{tbl:prior_models} shows that Llama-2-13b marginally outperforms humans on SARC2.0's pol-bal sarcasm detection task. We apply two input attribution interpretability methods: Integrated Gradients (IG) \citep{integrated_gradients}, and Shapley Additive Explanations (SHAP) \citep{shap} to the top performing model to attribute input features to a given classification. This analysis is performed on a subset of missed and correct predictions. Integrated Gradients, specifically Layer Integrated Gradients typically attributed classifications to uninformative tokens (e.g. BOS token, \textbackslash{n}, \textbackslash{n}\textbackslash{n\#\#\#}\textbackslash{n}). We take inspiration for our SHAP analysis from \citep{shapley_analysis_inspo}. Our SHAP values attributions were interpreted by one human. 

\subsubsection{Missed Classification Sample population vs. \textit{pol-bal} full population}
In an attempt to determine what the difference is between the full test set and the missed classification observations, we compared three different characteristics of each sample: token amount, post count, and named entity count (NER performed with BERT). Using a Mann-Whitney U test for the analysis of each attribute, we found that there is no statistical difference between the missed set and full test set observations with respect to the 3 attributes listed above.

\subsubsection{SHAP input attribution}

We recognize three categories of missed classifications. We took inspiration for this analysis from \cite{shapley_analysis_inspo}. 

\begin{itemize}
    \item \textbf{General Error}: A general error denotes a missed classification in which the model displayed confusion regarding or a misunderstanding of sarcasm. \Cref{fig:shap_general_1} presents the model classifying a sarcastic observation as non-sarcastic. The model attributes the reply's second word to the sarcastic label, however, the post text steers the model to classify the observation as non-sarcastic. The post text seemingly confused the model, and made it unable to detect the sarcasm in this observation. Interestingly, this sarcastic observation's non-sarcastic counterpart (seen in~\Cref{fig:shap_general_2}) was correctly classified by the model. This can be attributed to the non-sarcastic steering attributed to the observations' post text.
    
    \item \textbf{Contextual Error}: A contextual error occurs when the model cannot detect sarcasm because of an apparent lack of context. The model makes an apparent contextual error in~\Cref{fig:shap_context_1} when it determines the reply ``Cause the whole system is rigged'' is sarcastic. Without considering broad context, this reply would seem to sarcastically mock people who believe in election system rigging. However, there was significant belief in election rigging during the 2016 USA presidential election. The model may have forgot this contextual information during fine-tuning \citep{cata_interference}, did not realize said information, or did not realize that this observation was referencing the 2016 federal USA election during this observation's forward pass. Another example of contextual error can be seen in~\Cref{fig:shap_context_2} where the model does not recognize that the reply ``I'm officially an atheist now" is sarcastic given the contextual post ``Morgan Freeman endorses Clinton". The portion of the observation that mentions the actor contributes to the model outputting a sarcastic classification while the portion that mentions atheism contributes more to a non-sarcastic classification. This input attribution can be interpreted as the model not knowing, or not applying the knowledge that Morgan Freeman played the Christian God in the film, Bruce Almighty, which the observation's reply is most likely referencing.
    
    \item \textbf{Labeling Error}: \cite{khodak2017large} recognize the significant amount of false-negatives present in their \textit{entire} dataset (2.0\% false negatives vs 0.25\% true positives captured). This presence of false negatives is reflected in our error analysis. An example of a false negative in incorrectly classified observations can be seen in \Cref{fig:shap_fn_1} where the reply ``ummmm, awesome!" seems sarcastic, but is labeled as non-sarcastic in \textit{pol-bal}.
\end{itemize}

\subsection{Threats to validity}
\label{sec:threats}

We groups threats to validity into four categories (internal, construct, conclusion, and external) as presented in~\cite{yin2017case}.
 
\subsubsection{Internal validity}
In order to reduce the risk of errors, all experiments and data analysis were automated using Python scripts (which were cross-reviewed and validated by the authors).
 
\subsubsection{Construct validity}
This study uses accuracy and $F_1$-score to measure models' performance. Previous authors studying this dataset used the same metrics (see~\Cref{sec:sarc_pol_bal} for details).
 
In this study, several formulations of prompts were tested; the most effective ones were reported. Our prompts may be suboptimal, and we encourage the community to create better ones. The performance of the models may be further enhanced with such better prompts.
 
Furthermore, our research scope did not cover sophisticated prompt engineering techniques, such as $k$-shot~\citep{brown2020language}, chain of thought with reasoning~\citep{kojima2022large}, or retrieval augmented generation~\citep{lewis2020retrieval} that could improve the models' performance. However, this study aims not to achieve the best results for sarcasm detection but to examine how various attributes of models and prompts affect classification performance.
 
Another threat is that we use a regular expression to map the output of models to a label without looking at the output context. Thus, during zero-shot tests, regular expressions may misclassify specific outputs. We have sampled and eyeballed many outputs to mitigate this risk. We observed that the keyword ``yes" mapped to ``sarcastic" and the keyword ``no" mapped to ``non-sarcastic" in all cases. Furthermore, if neither label ``yes" nor ``no" was present, the output was meaningless and could not be categorized.

\subsubsection{Conclusion validity}
 
We do not know the exact content of the datasets used to train GPT models~\citep{brown2020language}. Therefore, it is possible that GPT models saw \textit{pol-bal} dataset during training. The fact that fine-tuning improves the results significantly over zero-shot experiments may suggest otherwise. At the very least, it may suggest that the models have ``forgotten'' this information to a significant extent.

\subsubsection{External validity}

Our study is based on a single dataset. Despite its popularity in sarcasm detection studies, our findings cannot be generalized to other datasets. However, the same empirical examination can also be applied to other datasets with well-designed and controlled experiments. Furthermore, this research serves as a case study that can help the community understand sarcasm detection trends and patterns and develop better LLMs to detect complex sentiments, such as sarcasm.
 
Although the Llama-2-13b fine-tuned model produced best-in-class prediction results, it does not imply that it will be used for those purposes in practice due to economic concerns. Due to the large number of parameters in LLMs, inference is expensive. The expense may be justified in business cases where sarcasm detection is critical.

\section{Conclusion and Next Steps}
\label{cha:conclusion}
This work on sarcasm detection is the first to explore how different attributes of Llama-2 and GPT affect their ability to detect sarcasm in the \textit{pol-bal} dataset. Fine-tuned Llama-2 and GPT outperform their zero-shot counterparts, while hyperparameter tuned Llama-2 models surpass larger, non-tuned GPT-3 models. We found that attributes like parameter count, and training epoch amount can affect a Llama-2 model's ability to detect sarcasm while attributes like version of a GPT may not have a significant effect on performance. This study affirms that the fine-tuned models tested and recently versioned zero-shot GPT-4 models are performant on NLU tasks, and gives researchers who wish to apply different LLMs to sarcasm detection, a methodological framework to do so. 
\subsection{Next Steps}
An interesting next step for this work is to better understand how the Llama-2-13b model performs sarcasm detection on the pol-bal dataset at a human level. This could be done by applying mechanistic interpretability methods like those seen in \cite{tigges2023linear} to determine which abstract concepts related to the \textit{pol-bal} task are represented in the fine-tuned model's learned parameters. 

Furthermore, open source models with parameter counts larger than 13 billion (e.g. Llama-2-70b) should be tested with our zero-shot and fine-tuned methodologies. Given the usage of LoRA in this research, different PEFT methods should also be tested. 

It would also be intriguing to apply a novel LoRA transferring technique (such as \textit{Trans-LoRA} \citep{wang2024textittransloradatafreetransferableparameter}) between multiple LLMs. 

Given the inherent political bias in the \textit{pol-bal} dataset, it would be compelling to apply instruction tuning to a model with the dataset in a way which aligns with AI safety. 

As discussed in \Cref{sec:threats}, a sarcasm classification model may not be very practical; however, a gated generative model like a mixture of experts which begins by determining whether an input is sarcastic could be useful in practice for a psychotherapist AI model or a similarly emotionally intelligent chat-bot.

\bibliography{sn-bibliography}

\begin{appendices}

\section{Pairwise GPT McNemar tests}

\subsection{Results of the pairwise McNemar's test for fine-tuned GPT models}
\label{sec:mcnemar_fine_tuned}
McNemar's test~\citep{mcnemar1947note} is applied to predictions of each pair of classification models as per~\cite{dietterich1998approximate} to compare the performance of different-sized fine-tuned GPT-3 models as well as differently versioned GPT models.~\Cref{results-mcnemar-ft-noenv} shows the results of the pairwise comparison of the models included in RQ1 while~\Cref{results-mcnemar-ft-version} shows the results for RQ4. 

\section{Results of the pairwise McNemar's test for different versions of GPT-3.5-turbo and GPT-4}
\label{sec:mcnemar_versions}
\begin{figure}[H]
\centering
\includegraphics[width=0.6\textwidth]{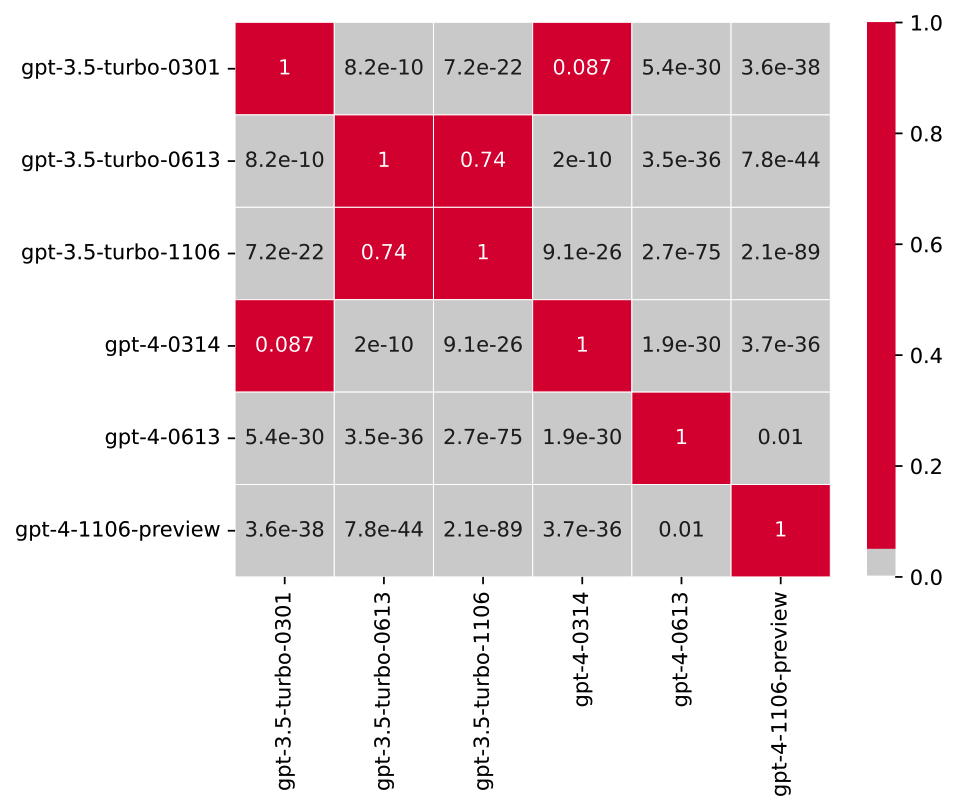}
\caption{Pairwise McNemar's test of ChatGPT models with logit bias.}
\label{results-mcnemar-zeroshot-chat-nodomain-logit}
\end{figure}

In this section, we compare the difference in performance for different versions of GPT-3.5-turbo and GPT-4 models for the zero-shot case (with and without logit bias). McNemar's test~\citep{mcnemar1947note} is applied to predictions of each pair of classification models as per~\cite{dietterich1998approximate}.~\Cref{results-mcnemar-zeroshot-chat-nodomain-nologit} shows the pairwise comparison results of the models tested without logit bias.~\Cref{results-mcnemar-zeroshot-chat-nodomain-logit} shows the pairwise comparison results of the models tested with logit bias.

\section{Llama-2 Results}

\begin{figure}[H]
\centering
\includegraphics[width=0.6\textwidth]{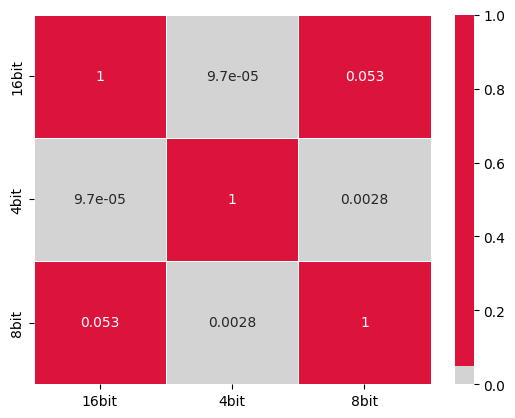}
\caption{Pairwise McNemar's test of zero-shot Llama-2-chat 7 billion parameter models.}
\label{fig:mcnemar-llama-chat-7b}
\end{figure}

\subsection{Results of the pairwise McNemar's test for zero-shot Llama-2-chat models}
\begin{figure}[H]
\centering
\includegraphics[width=0.6\textwidth]{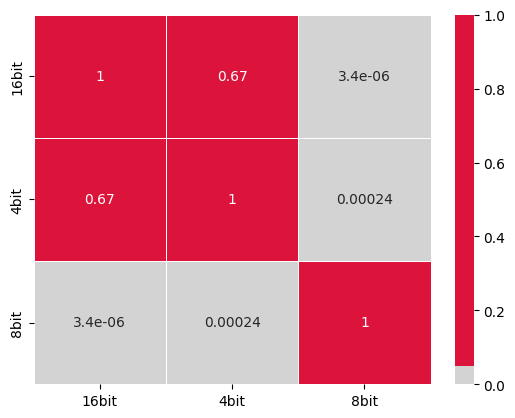}
\caption{Pairwise McNemar's test of zero-shot Llama-2-chat 13 billion parameter models.}
\label{fig:mcnemar-llama-chat-13b}
\end{figure}

\subsection{Pairwise Hyperparameter Wilcoxon Signed Rank Tests}
\label{sec:ranktests}
\begin{figure}[H]
\label{fig:utest_quant}
\centering
\centering
\includegraphics[width=1\textwidth]{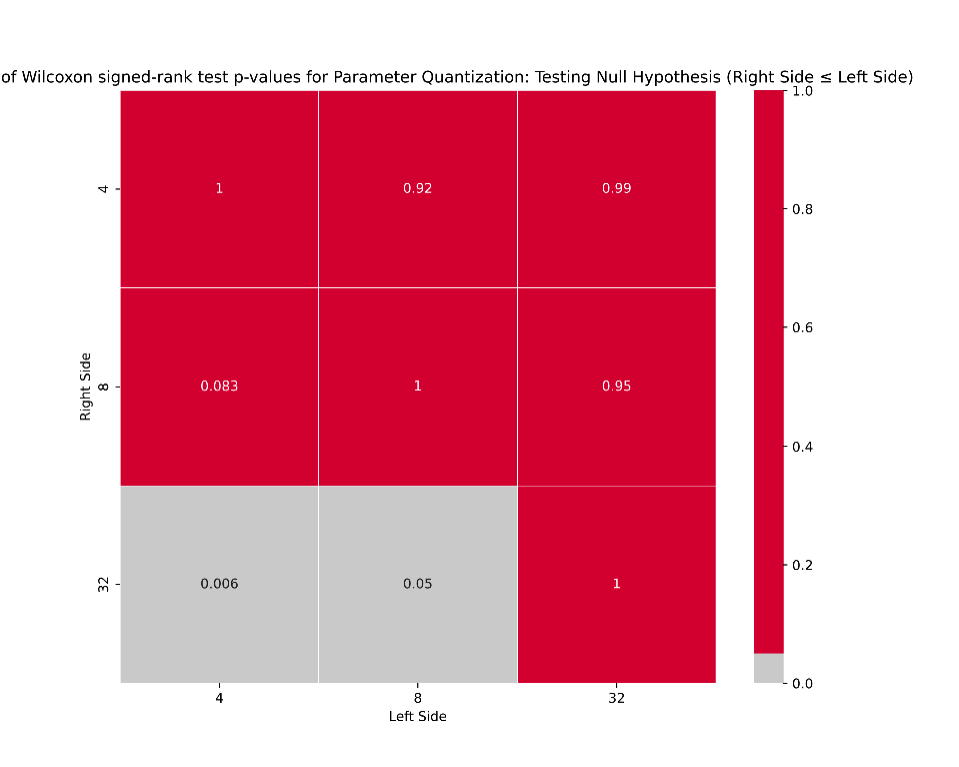}
\caption{Pairwise Wilcoxon signed-rank test of parameter precision hyperparameter values.}
\end{figure}

\begin{figure}[H]
\label{fig:utest_batch_size}
\centering
\centering
\includegraphics[width=1\textwidth]{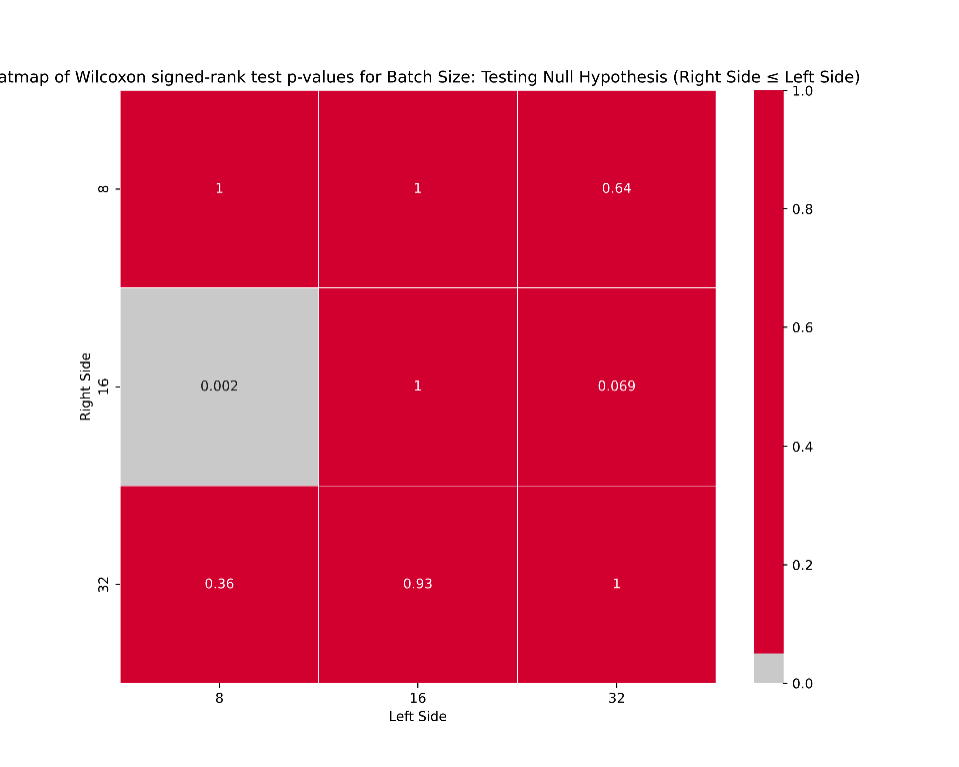}
\caption{Pairwise Wilcoxon signed-rank test of batch size hyperparameter values.}
\end{figure}

\begin{figure}[H]
\label{fig:utest_epoch_count}
\centering
\centering
\includegraphics[width=1\textwidth]{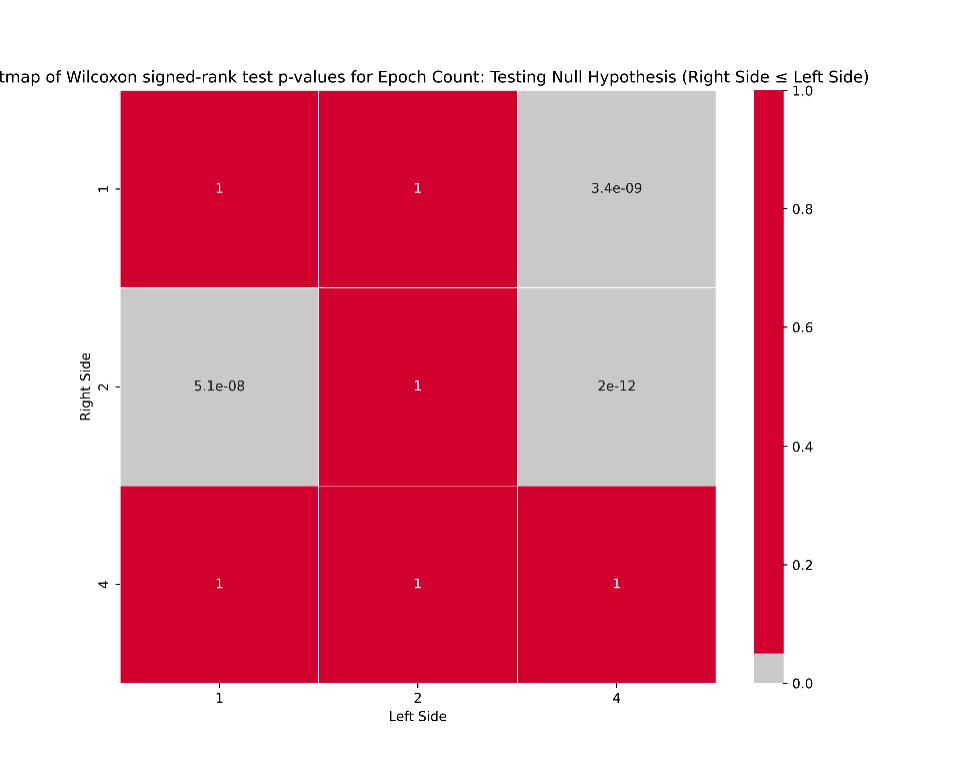}
\caption{Pairwise Wilcoxon signed-rank test of epoch count hyperparameter values.}
\end{figure}

\begin{figure}[H]
\label{fig:utest_lora_rank}
\centering
\centering
\includegraphics[width=1\textwidth]{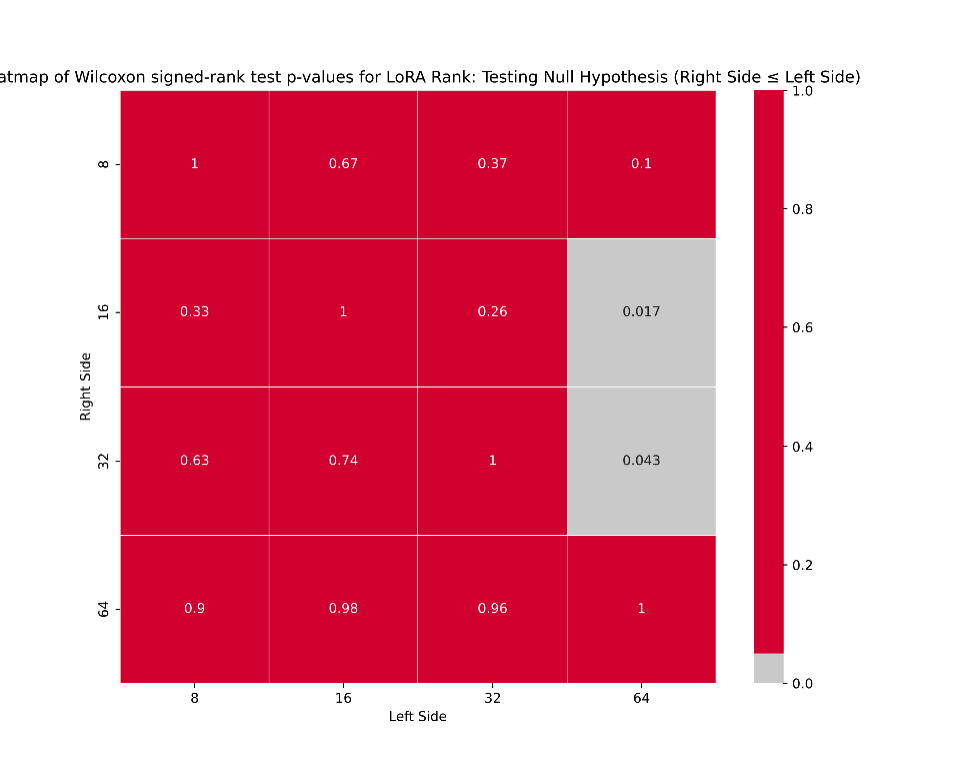}
\caption{Pairwise Wilcoxon signed-rank test of LoRA rank hyperparameter values.}
\end{figure}

\subsection{Llama fine-tuned Shapley Error Analysis}
The contribution of each token within an input sequence can be either positive or negative with respect to the output one a label (``no" denoted as LABEL\textunderscore0 or ``yes" denoted as LABEL\textunderscore1). The contributions are noted with respect to the expected value of an input, represented as the most red label with a black underline. A token's contribution to a label is correlated to the intensity of its colour; specifically, the more intensely red a token is highlighted, the more it contributed to the selected label (always the expected value in this case), while the more intensely blue a token is highlighted, the more it contributed to the opposite label, and the more transparent the highlighting of a token, the more neutral it was in contributing to a certain output label.
\begin{figure}[H]
    \centering
    \includegraphics[width=0.75\linewidth]{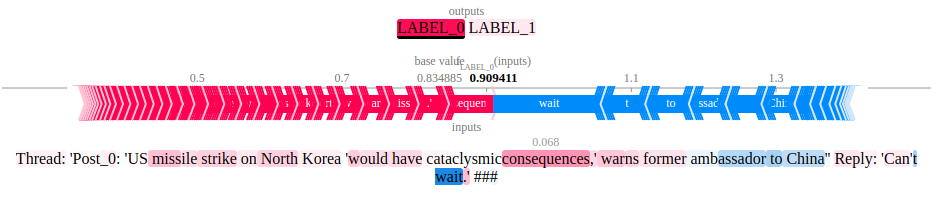}
    \caption{General Error example with a sarcastic observation.}
    \label{fig:shap_general_1}
\end{figure}

\begin{figure}[H]
    \centering
    \includegraphics[width=0.75\linewidth]{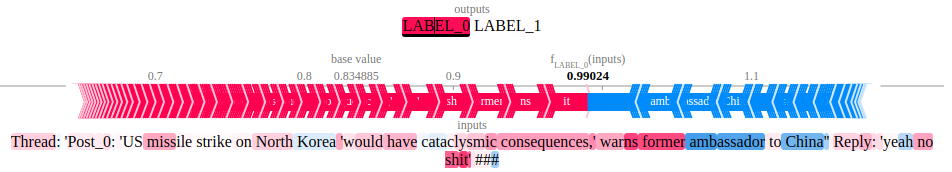}
    \caption{General Error example's non-sarcastic counterpart.}
    \label{fig:shap_general_2}
\end{figure}

\begin{figure}[H]
    \centering
    \includegraphics[width=0.75\linewidth]{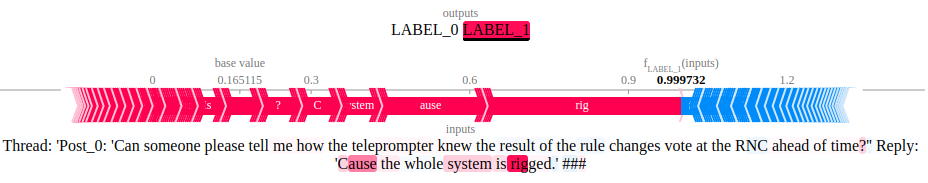}
    \caption{Contextual Error example with a non-sarcastic observation.}
    \label{fig:shap_context_1}
\end{figure}

\begin{figure}[H]
    \centering
    \includegraphics[width=0.75\linewidth]{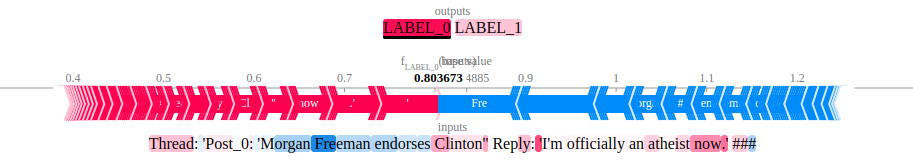}
    \caption{Contextual Error example with a sarcastic observation.}
    \label{fig:shap_context_2}
\end{figure}

\begin{figure}[H]
    \centering
    \includegraphics[width=0.75\linewidth]{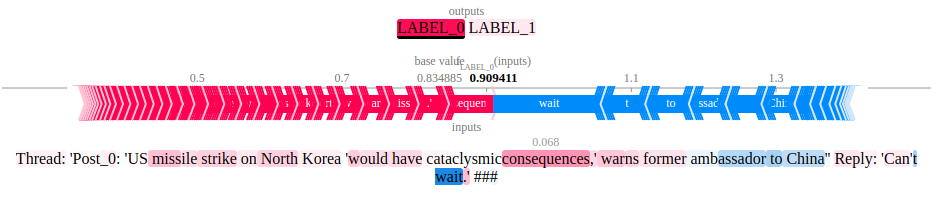}
    \caption{False Negative example with a non-sarcastic labeled observation.}
    \label{fig:shap_fn_1}
\end{figure}

\subsection{Llama fine-tuned results}
We performed a grid search on hyperparameter values when fine-tuning the Llama models shown in \Cref{tab:hyperparameters}. We examined 216 combinations of hyperparameters; \Cref{tbl:llama_full_ft_results} holds these data.
\begin{longtable}{@{}rrrrrrrr@{}}
\caption{Performance metrics for hyperparameter sweep values from \Cref{tab:hyperparameters} on Llama models. Sorted in descending order by accuracy.\label{tbl:llama_full_ft_results}}\\
\toprule
Accuracy & $F_1$  & MCC & LoRA  & Batch  & Train  & Model  & Parameter \\
 &  Score & & Rank &  Size &  Epochs &  Size & Precision \\\midrule
\endhead 
\hline
\multicolumn{8}{@{}r@{}}{continued \ldots}\\
\endfoot
\hline
\endlastfoot

0.835 & 0.834 & 0.669 & 16 & 8 & 2 & 13 & 32 \\
0.832 & 0.835 & 0.665 & 32 & 16 & 2 & 13 & 32 \\
0.832 & 0.835 & 0.665 & 32 & 32 & 2 & 13 & 4 \\
0.832 & 0.834 & 0.664 & 64 & 16 & 2 & 13 & 32 \\
0.831 & 0.833 & 0.663 & 64 & 32 & 2 & 13 & 32 \\
0.831 & 0.834 & 0.663 & 16 & 32 & 2 & 13 & 4 \\
0.831 & 0.834 & 0.663 & 8 & 32 & 2 & 13 & 8 \\
0.831 & 0.834 & 0.662 & 16 & 8 & 1 & 13 & 32 \\
0.831 & 0.832 & 0.661 & 16 & 16 & 2 & 13 & 32 \\
0.831 & 0.833 & 0.662 & 32 & 32 & 2 & 13 & 32 \\
0.830 & 0.831 & 0.661 & 16 & 16 & 2 & 13 & 4 \\
0.830 & 0.832 & 0.661 & 64 & 16 & 2 & 13 & 4 \\
0.830 & 0.831 & 0.661 & 64 & 32 & 2 & 13 & 4 \\
0.830 & 0.831 & 0.660 & 8 & 8 & 1 & 13 & 4 \\
0.830 & 0.830 & 0.660 & 8 & 16 & 2 & 13 & 32 \\
0.830 & 0.830 & 0.659 & 32 & 16 & 2 & 13 & 8 \\
0.830 & 0.833 & 0.660 & 16 & 16 & 1 & 13 & 8 \\
0.830 & 0.830 & 0.659 & 8 & 8 & 2 & 13 & 8 \\
0.829 & 0.832 & 0.659 & 8 & 16 & 2 & 13 & 8 \\
0.829 & 0.830 & 0.658 & 8 & 32 & 2 & 13 & 32 \\
0.829 & 0.827 & 0.658 & 16 & 32 & 2 & 13 & 32 \\
0.829 & 0.830 & 0.658 & 16 & 8 & 2 & 13 & 8 \\
0.829 & 0.833 & 0.659 & 8 & 16 & 1 & 13 & 8 \\
0.829 & 0.829 & 0.658 & 16 & 32 & 2 & 13 & 8 \\
0.829 & 0.829 & 0.657 & 16 & 16 & 2 & 13 & 8 \\
0.829 & 0.832 & 0.658 & 32 & 16 & 2 & 7 & 32 \\
0.828 & 0.830 & 0.657 & 64 & 32 & 2 & 13 & 8 \\
0.828 & 0.829 & 0.657 & 8 & 32 & 2 & 7 & 8 \\
0.828 & 0.830 & 0.656 & 32 & 8 & 1 & 13 & 4 \\
0.828 & 0.831 & 0.656 & 8 & 16 & 1 & 13 & 4 \\
0.828 & 0.831 & 0.656 & 16 & 32 & 1 & 13 & 8 \\
0.828 & 0.830 & 0.656 & 32 & 16 & 1 & 13 & 32 \\
0.827 & 0.831 & 0.655 & 16 & 16 & 1 & 13 & 32 \\
0.827 & 0.828 & 0.655 & 8 & 16 & 2 & 7 & 8 \\
0.827 & 0.830 & 0.654 & 32 & 8 & 1 & 13 & 32 \\
0.827 & 0.831 & 0.655 & 16 & 16 & 1 & 13 & 4 \\
0.827 & 0.827 & 0.654 & 8 & 8 & 2 & 13 & 4 \\
0.827 & 0.827 & 0.654 & 8 & 8 & 2 & 13 & 32 \\
0.827 & 0.827 & 0.654 & 64 & 32 & 2 & 7 & 8 \\
0.826 & 0.826 & 0.653 & 32 & 16 & 2 & 13 & 4 \\
0.826 & 0.829 & 0.653 & 32 & 32 & 1 & 13 & 4 \\
0.826 & 0.828 & 0.652 & 64 & 8 & 2 & 13 & 4 \\
0.826 & 0.828 & 0.652 & 32 & 16 & 1 & 13 & 4 \\
0.826 & 0.824 & 0.651 & 8 & 16 & 2 & 13 & 4 \\
0.826 & 0.826 & 0.651 & 8 & 32 & 2 & 13 & 4 \\
0.825 & 0.828 & 0.651 & 8 & 8 & 1 & 13 & 8 \\
0.825 & 0.829 & 0.651 & 16 & 16 & 2 & 7 & 32 \\
0.825 & 0.826 & 0.651 & 32 & 8 & 2 & 13 & 4 \\
0.825 & 0.825 & 0.65 & 32 & 8 & 2 & 13 & 32 \\
0.825 & 0.824 & 0.65 & 32 & 32 & 2 & 13 & 8 \\
0.825 & 0.825 & 0.649 & 16 & 32 & 2 & 7 & 32 \\
0.825 & 0.824 & 0.649 & 32 & 8 & 2 & 13 & 8 \\
0.825 & 0.827 & 0.650 & 64 & 16 & 1 & 13 & 4 \\
0.824 & 0.827 & 0.649 & 16 & 8 & 1 & 13 & 8 \\
0.824 & 0.829 & 0.650 & 64 & 32 & 1 & 7 & 32 \\
0.824 & 0.830 & 0.650 & 64 & 16 & 1 & 13 & 32 \\
0.824 & 0.823 & 0.649 & 8 & 8 & 2 & 7 & 32 \\
0.824 & 0.825 & 0.648 & 8 & 8 & 1 & 7 & 32 \\
0.824 & 0.826 & 0.648 & 8 & 16 & 2 & 7 & 32 \\
0.824 & 0.824 & 0.648 & 16 & 32 & 2 & 7 & 4 \\
0.824 & 0.829 & 0.649 & 64 & 32 & 1 & 13 & 32 \\
0.824 & 0.825 & 0.648 & 32 & 16 & 1 & 13 & 8 \\
0.824 & 0.823 & 0.648 & 64 & 8 & 2 & 13 & 8 \\
0.824 & 0.829 & 0.649 & 8 & 32 & 1 & 13 & 4 \\
0.824 & 0.825 & 0.648 & 32 & 32 & 2 & 7 & 8 \\
0.824 & 0.827 & 0.648 & 64 & 8 & 1 & 13 & 8 \\
0.823 & 0.824 & 0.647 & 16 & 8 & 2 & 7 & 32 \\
0.823 & 0.828 & 0.648 & 64 & 16 & 1 & 13 & 8 \\
0.823 & 0.827 & 0.647 & 32 & 32 & 1 & 13 & 8 \\
0.823 & 0.829 & 0.648 & 16 & 32 & 1 & 13 & 32 \\
0.823 & 0.826 & 0.646 & 64 & 16 & 1 & 7 & 4 \\
0.823 & 0.824 & 0.646 & 64 & 16 & 2 & 13 & 8 \\
0.823 & 0.827 & 0.647 & 8 & 32 & 1 & 13 & 32 \\
0.823 & 0.828 & 0.647 & 16 & 16 & 1 & 7 & 32 \\
0.823 & 0.826 & 0.646 & 16 & 8 & 1 & 13 & 4 \\
0.822 & 0.825 & 0.645 & 16 & 8 & 2 & 7 & 4 \\
0.822 & 0.823 & 0.645 & 32 & 32 & 2 & 7 & 32 \\
0.822 & 0.821 & 0.644 & 8 & 32 & 2 & 7 & 4 \\
0.822 & 0.825 & 0.645 & 8 & 32 & 4 & 13 & 32 \\
0.822 & 0.824 & 0.644 & 16 & 16 & 2 & 7 & 4 \\
0.822 & 0.823 & 0.644 & 8 & 32 & 2 & 7 & 32 \\
0.822 & 0.824 & 0.644 & 64 & 32 & 4 & 13 & 4 \\
0.821 & 0.822 & 0.643 & 8 & 16 & 2 & 7 & 4 \\
0.821 & 0.823 & 0.643 & 64 & 8 & 1 & 13 & 32 \\
0.821 & 0.820 & 0.643 & 16 & 8 & 2 & 13 & 4 \\
0.821 & 0.825 & 0.643 & 64 & 16 & 4 & 13 & 8 \\
0.821 & 0.822 & 0.643 & 64 & 8 & 2 & 7 & 4 \\
0.821 & 0.825 & 0.643 & 64 & 32 & 1 & 13 & 8 \\
0.821 & 0.825 & 0.643 & 64 & 16 & 1 & 7 & 32 \\
0.821 & 0.821 & 0.642 & 32 & 8 & 2 & 7 & 4 \\
0.821 & 0.819 & 0.642 & 32 & 16 & 2 & 7 & 8 \\
0.821 & 0.822 & 0.642 & 16 & 16 & 2 & 7 & 8 \\
0.821 & 0.820 & 0.642 & 64 & 16 & 2 & 7 & 4 \\
0.821 & 0.820 & 0.642 & 64 & 32 & 2 & 7 & 32 \\
0.820 & 0.820 & 0.641 & 8 & 16 & 4 & 13 & 4 \\
0.820 & 0.822 & 0.641 & 8 & 8 & 1 & 7 & 8 \\
0.820 & 0.819 & 0.641 & 8 & 8 & 1 & 7 & 4 \\
0.820 & 0.821 & 0.641 & 64 & 32 & 4 & 13 & 8 \\
0.820 & 0.820 & 0.641 & 16 & 8 & 2 & 7 & 8 \\
0.820 & 0.818 & 0.641 & 32 & 8 & 1 & 7 & 32 \\
0.820 & 0.820 & 0.641 & 8 & 32 & 4 & 7 & 8 \\
0.820 & 0.823 & 0.641 & 64 & 8 & 1 & 13 & 4 \\
0.820 & 0.820 & 0.640 & 8 & 8 & 2 & 7 & 8 \\
0.820 & 0.819 & 0.639 & 32 & 32 & 4 & 13 & 4 \\
0.820 & 0.824 & 0.640 & 8 & 16 & 1 & 7 & 8 \\
0.820 & 0.822 & 0.640 & 32 & 8 & 1 & 7 & 4 \\
0.820 & 0.821 & 0.640 & 64 & 16 & 2 & 7 & 32 \\
0.820 & 0.824 & 0.640 & 8 & 32 & 1 & 7 & 8 \\
0.819 & 0.822 & 0.639 & 32 & 16 & 1 & 7 & 8 \\
0.819 & 0.820 & 0.639 & 8 & 16 & 4 & 7 & 32 \\
0.819 & 0.824 & 0.640 & 32 & 32 & 1 & 13 & 32 \\
0.819 & 0.827 & 0.641 & 16 & 32 & 1 & 13 & 4 \\
0.819 & 0.819 & 0.638 & 32 & 16 & 1 & 7 & 4 \\
0.819 & 0.823 & 0.639 & 16 & 8 & 4 & 13 & 8 \\
0.819 & 0.820 & 0.638 & 32 & 16 & 2 & 7 & 4 \\
0.819 & 0.822 & 0.638 & 64 & 32 & 1 & 13 & 4 \\
0.819 & 0.821 & 0.638 & 16 & 32 & 4 & 13 & 8 \\
0.819 & 0.820 & 0.637 & 16 & 8 & 4 & 13 & 32 \\
0.819 & 0.820 & 0.637 & 32 & 16 & 4 & 7 & 4 \\
0.819 & 0.818 & 0.637 & 32 & 8 & 1 & 13 & 8 \\
0.818 & 0.822 & 0.637 & 8 & 8 & 4 & 13 & 32 \\
0.818 & 0.820 & 0.637 & 32 & 8 & 2 & 7 & 8 \\
0.818 & 0.822 & 0.637 & 16 & 8 & 1 & 7 & 8 \\
0.818 & 0.819 & 0.637 & 32 & 32 & 2 & 7 & 4 \\
0.818 & 0.819 & 0.636 & 32 & 16 & 4 & 13 & 32 \\
0.818 & 0.819 & 0.636 & 16 & 16 & 4 & 13 & 8 \\
0.818 & 0.825 & 0.638 & 8 & 32 & 1 & 13 & 8 \\
0.818 & 0.817 & 0.636 & 8 & 8 & 2 & 7 & 4 \\
0.818 & 0.815 & 0.636 & 16 & 32 & 2 & 7 & 8 \\
0.818 & 0.823 & 0.636 & 16 & 32 & 1 & 7 & 32 \\
0.818 & 0.820 & 0.636 & 32 & 32 & 4 & 13 & 8 \\
0.818 & 0.818 & 0.635 & 32 & 8 & 1 & 7 & 8 \\
0.817 & 0.818 & 0.635 & 16 & 16 & 1 & 7 & 8 \\
0.817 & 0.820 & 0.635 & 32 & 8 & 4 & 13 & 4 \\
0.817 & 0.814 & 0.635 & 64 & 32 & 4 & 13 & 32 \\
0.817 & 0.822 & 0.636 & 32 & 16 & 1 & 7 & 32 \\
0.817 & 0.820 & 0.635 & 16 & 16 & 4 & 13 & 4 \\
0.817 & 0.824 & 0.636 & 32 & 32 & 1 & 7 & 4 \\
0.817 & 0.816 & 0.634 & 64 & 16 & 4 & 13 & 4 \\
0.817 & 0.819 & 0.634 & 32 & 32 & 4 & 13 & 32 \\
0.817 & 0.822 & 0.635 & 16 & 16 & 1 & 7 & 4 \\
0.817 & 0.818 & 0.634 & 32 & 32 & 4 & 7 & 8 \\
0.817 & 0.819 & 0.634 & 16 & 32 & 4 & 7 & 8 \\
0.817 & 0.824 & 0.636 & 8 & 32 & 1 & 7 & 32 \\
0.817 & 0.821 & 0.634 & 32 & 32 & 4 & 7 & 32 \\
0.817 & 0.819 & 0.633 & 64 & 16 & 4 & 13 & 32 \\
0.817 & 0.818 & 0.633 & 8 & 16 & 1 & 13 & 32 \\
0.816 & 0.822 & 0.634 & 8 & 8 & 4 & 13 & 4 \\
0.816 & 0.817 & 0.632 & 16 & 32 & 4 & 7 & 4 \\
0.816 & 0.820 & 0.633 & 64 & 32 & 1 & 7 & 4 \\
0.816 & 0.816 & 0.632 & 32 & 8 & 2 & 7 & 32 \\
0.816 & 0.817 & 0.632 & 32 & 16 & 4 & 13 & 4 \\
0.816 & 0.814 & 0.631 & 32 & 16 & 4 & 7 & 8 \\
0.815 & 0.815 & 0.631 & 8 & 8 & 4 & 7 & 32 \\
0.815 & 0.818 & 0.631 & 16 & 32 & 4 & 13 & 32 \\
0.815 & 0.819 & 0.631 & 64 & 8 & 1 & 7 & 8 \\
0.815 & 0.821 & 0.632 & 64 & 32 & 4 & 7 & 4 \\
0.815 & 0.814 & 0.630 & 16 & 8 & 1 & 7 & 4 \\
0.815 & 0.819 & 0.631 & 32 & 8 & 4 & 13 & 32 \\
0.815 & 0.816 & 0.630 & 16 & 8 & 4 & 7 & 32 \\
0.815 & 0.816 & 0.630 & 64 & 16 & 4 & 7 & 32 \\
0.815 & 0.813 & 0.630 & 16 & 16 & 4 & 13 & 32 \\
0.815 & 0.820 & 0.631 & 16 & 8 & 4 & 13 & 4 \\
0.815 & 0.818 & 0.630 & 8 & 32 & 4 & 13 & 8 \\
0.814 & 0.817 & 0.629 & 64 & 32 & 4 & 7 & 8 \\
0.814 & 0.814 & 0.629 & 64 & 32 & 2 & 7 & 4 \\
0.814 & 0.820 & 0.630 & 64 & 32 & 4 & 7 & 32 \\
0.814 & 0.819 & 0.630 & 8 & 16 & 4 & 13 & 8 \\
0.814 & 0.821 & 0.630 & 32 & 32 & 1 & 7 & 8 \\
0.814 & 0.816 & 0.628 & 8 & 16 & 1 & 7 & 4 \\
0.814 & 0.816 & 0.627 & 64 & 16 & 4 & 7 & 8 \\
0.813 & 0.815 & 0.627 & 16 & 16 & 4 & 7 & 32 \\
0.813 & 0.821 & 0.629 & 32 & 32 & 1 & 7 & 32 \\
0.813 & 0.815 & 0.626 & 8 & 32 & 4 & 13 & 4 \\
0.813 & 0.813 & 0.626 & 16 & 32 & 4 & 7 & 32 \\
0.813 & 0.815 & 0.626 & 8 & 8 & 4 & 13 & 8 \\
0.812 & 0.816 & 0.625 & 16 & 16 & 4 & 7 & 8 \\
0.812 & 0.815 & 0.625 & 64 & 8 & 4 & 13 & 8 \\
0.812 & 0.817 & 0.626 & 16 & 32 & 1 & 7 & 8 \\
0.812 & 0.812 & 0.624 & 32 & 16 & 4 & 7 & 32 \\
0.812 & 0.817 & 0.625 & 64 & 16 & 1 & 7 & 8 \\
0.811 & 0.814 & 0.623 & 32 & 32 & 4 & 7 & 4 \\
0.811 & 0.817 & 0.624 & 64 & 32 & 1 & 7 & 8 \\
0.811 & 0.812 & 0.621 & 64 & 8 & 4 & 7 & 32 \\
0.810 & 0.809 & 0.620 & 16 & 16 & 4 & 7 & 4 \\
0.809 & 0.811 & 0.619 & 8 & 32 & 4 & 7 & 4 \\
0.809 & 0.813 & 0.619 & 32 & 16 & 4 & 13 & 8 \\
0.808 & 0.808 & 0.615 & 64 & 8 & 4 & 7 & 4 \\
0.807 & 0.809 & 0.615 & 8 & 16 & 4 & 7 & 8 \\
0.807 & 0.811 & 0.614 & 8 & 16 & 4 & 13 & 32 \\
0.807 & 0.810 & 0.613 & 16 & 32 & 4 & 13 & 4 \\
0.806 & 0.808 & 0.613 & 8 & 16 & 4 & 7 & 4 \\
0.806 & 0.808 & 0.612 & 64 & 8 & 4 & 13 & 4 \\
0.806 & 0.806 & 0.612 & 16 & 8 & 4 & 7 & 8 \\
0.805 & 0.810 & 0.611 & 32 & 8 & 4 & 7 & 32 \\
0.805 & 0.805 & 0.61 & 64 & 8 & 1 & 7 & 4 \\
0.805 & 0.802 & 0.61 & 64 & 16 & 4 & 7 & 4 \\
0.805 & 0.814 & 0.612 & 8 & 32 & 1 & 7 & 4 \\
0.804 & 0.805 & 0.609 & 8 & 8 & 4 & 7 & 4 \\
0.803 & 0.806 & 0.607 & 16 & 32 & 1 & 7 & 4 \\
0.803 & 0.800 & 0.607 & 32 & 8 & 4 & 7 & 8 \\
0.802 & 0.804 & 0.604 & 8 & 32 & 4 & 7 & 32 \\
0.789 & 0.773 & 0.583 & 8 & 8 & 1 & 13 & 32 \\
0.648 & 0.610 & 0.302 & 64 & 8 & 2 & 7 & 32 \\
0.631 & 0.617 & 0.262 & 32 & 8 & 4 & 7 & 4 \\
0.615 & 0.561 & 0.236 & 8 & 8 & 4 & 7 & 8 \\
0.607 & 0.505 & 0.235 & 64 & 8 & 4 & 13 & 32 \\
0.585 & 0.388 & 0.222 & 64 & 8 & 2 & 7 & 8 \\
0.582 & 0.646 & 0.176 & 64 & 8 & 2 & 13 & 32 \\
0.519 & 0.273 & 0.052 & 64 & 8 & 1 & 7 & 32 \\
0.501 & 0.436 & 0.001 & 16 & 8 & 1 & 7 & 32 \\
0.500 & 0.667 & 0.000 & 8 & 16 & 1 & 7 & 32 \\
0.500 & 0.000 & 0.000 & 32 & 8 & 4 & 13 & 8 \\
0.500 & 0.000 & 0.000 & 16 & 8 & 4 & 7 & 4 \\
0.500 & 0.001 & 0.000 & 64 & 8 & 4 & 7 & 8 \\
0.499 & 0.657 & -0.004 & 64 & 16 & 2 & 7 & 8 \\ 
\end{longtable}

\end{appendices}

\end{document}